\documentclass{article}

 \usepackage[preprint]{neurips_2025}

\usepackage[utf8]{inputenc} 
\usepackage[T1]{fontenc}    
\usepackage{hyperref}       
\usepackage{url}            
\usepackage{booktabs}       
\usepackage{amsfonts}       
\usepackage{nicefrac}       
\usepackage{microtype}      
\usepackage{xcolor}         
\usepackage{enumitem}

\usepackage{amsmath,amsfonts,bm}
\usepackage{nicefrac}

\def\eqref#1{equation~\ref{#1}}

\def\1{\bm{1}}

\def\vx{{\bm{x}}}

\DeclareMathAlphabet{\mathsfit}{\encodingdefault}{\sfdefault}{m}{sl}
\SetMathAlphabet{\mathsfit}{bold}{\encodingdefault}{\sfdefault}{bx}{n}

\newcommand{\be}{\begin{eqnarray} \begin{aligned}}
\newcommand{\ee}{\end{aligned} \end{eqnarray} }
\newcommand{\benn}{\begin{eqnarray*} \begin{aligned}}
\newcommand{\eenn}{\end{aligned} \end{eqnarray*} }
\newcommand{\xhat}{\hat{\bm x}}

\newcommand{\snr}{\gamma}

\newcommand{\csig}{\sqrt{\gamma/(1+\gamma)}}
\newcommand{\cnoise}{\sqrt{1/(1+\gamma)}}

\newcommand{\vz}{\bm z_\snr}

\newcommand{\half}{\nicefrac{1}{2}}
\newcommand{\eps}{\bm \epsilon}

\DeclareMathOperator{\mmse}{mmse}

\usepackage{hyperref}
\usepackage{url}
\usepackage{empheq} 
\usepackage{booktabs}  
\usepackage{color}
\usepackage{xcolor}  
\usepackage{colortbl}
\usepackage{multirow}
\usepackage{longtable}
\usepackage{float} 
\usepackage{graphicx}
\usepackage{booktabs}
\usepackage{colortbl}
\usepackage{multirow}
\usepackage{longtable}
\usepackage{lscape}
\usepackage{rotating}
\usepackage{subcaption}
\usepackage{amsthm}
\usepackage{glossaries}

\newacronym{mind}{MIND}{Mutual Information via Neural Diffusion}
\newacronym{minde}{MINDE}{Mutual Information Neural Diffusion Estimator}

\definecolor{longxuancolor}{HTML}{DC143C}

\title{MMG: Mutual Information Estimation via the MMSE Gap in Diffusion}

\author{%
  \begin{tabular}{c}
    Longxuan Yu\textsuperscript{1}\thanks{Equal contribution.} \qquad
    Xing Shi\textsuperscript{1}\footnotemark[1] \qquad
    Xianghao Kong\textsuperscript{1} \qquad
    Tong Jia\textsuperscript{2} \qquad
    Greg Ver Steeg\textsuperscript{1}
    \\[0.5em]
    \textsuperscript{1}University of California, Riverside \qquad
    \textsuperscript{2}University of California, San Diego
    \\[0.5em]
    \texttt{\{ylong030, xshi068, xkong016, gregoryv\}@ucr.edu}, \quad
    \texttt{tojia@ucsd.edu}
  \end{tabular}
}

\begin{document}

\maketitle

\begin{abstract}
  Mutual information (MI) is one of the most general ways to measure relationships between random variables, but estimating this quantity for complex systems is challenging. 
Denoising diffusion models have recently set a new bar for density estimation, so it is natural to consider whether these methods could also be used to improve MI estimation. 
Using the recently introduced information-theoretic formulation of denoising diffusion models, we show the diffusion models can be used in a straightforward way to estimate MI. 
In particular, the MI corresponds to half the gap in the Minimum Mean Square Error (MMSE) between conditional and unconditional diffusion, integrated over all Signal-to-Noise-Ratios (SNRs) in the noising process. 
Our approach not only passes self-consistency tests but also outperforms traditional and score-based diffusion MI estimators. Furthermore, our method leverages  adaptive importance sampling to achieve scalable MI estimation, while maintaining strong performance even when the MI is high. Code to reproduce experiments is at
\url{https://github.com/fengsxy/Diffusion_MI},
and the unified MI library is at
\url{https://github.com/fengsxy/Diffusion-MI}.
\end{abstract}

\section{Introduction}
Estimating Mutual Information (MI) from samples is a fundamental problem with widespread applications.  
Methods based on local density estimation \citep{kraskov2004, pal2010estimation, gao2015efficient} have been displaced by variational approaches using neural networks \citep{poole2019variational} to estimate lower bounds on MI \citep{belghazi2018mutual, nguyen2010estimating}. 
Unfortunately, these approaches may have sample complexity or variance which scale exponentially with the true MI \citep{gao2015efficient, mcallester2020formal, oord2018representation}. In practical scenarios, the MI between two variables is usually unknown, making reliable estimation challenging. To tackle this, \citet{song2019understanding} introduced three self-consistency experiments designed to evaluate the robustness of MI estimators. For a more standardized and comprehensive evaluation, \citet{czyż2023normalevaluationmutualinformation} developed a benchmark consisting of 40 synthetic datasets derived from diverse distributions, providing a consistent basis for assessing the performance of different MI estimators.

Denoising diffusion models~\cite{jaschaneq,ddpm,vdm,itd} have dramatically improved the modeling of complex distributions, igniting a new industry for AI art generation~\cite{latent_diffusion,dalle2,imagen}. 
Recently, \citet{minde} introduced a MI estimator, MINDE, based on score-based diffusion models. This estimator not only achieved an 87.5\% estimation success rate on the benchmark of \citet{czyż2023normalevaluationmutualinformation} but also passed the self-consistency tests. However, its reliance on accurately approximating the log-density gradient can be a challenging intermediate step. 
This motivates a more direct formulation, connecting MI to the denoising objective itself rather than its gradient.

In this paper, we exploit the recently discovered connections between information theory and diffusion models to derive an elegant and effective approach to MI estimation~\citep{guo,itd}. Our contributions are as follows:

\begin{itemize}
    \item We show that conditional density estimation and mutual information estimation can both be written exactly in terms of the global optimum of a denoising objective.
    Conditional density estimation corresponds to a gap between MMSEs~\citep{itd} for conditional and unconditional denoising diffusion models, and mutual information is the expected gap over all data, leading to the  \textbf{M}utual Information Estimation via the \textbf{M}MSE \textbf{G}ap in Diffusion (MMG) estimator. 
    \item We develop an adaptive importance sampling scheme that tailors the integration to the specific data distribution. By dynamically fitting a sampling distribution to the MMSE gap, this technique significantly improves the precision and efficiency of the final MI estimate.
    \item MMG has passed all self-consistency tests and achieved state-of-the-art results across multiple tasks, particularly excelling in high MI estimation, and has outperformed the current leading estimator, MINDE, in these scenarios. 
    \item We release a unified PyTorch library that, for the first time, brings diffusion-based and established neural MI estimators into a single, consistent framework. Its simple API is designed to streamline their future side-by-side evaluation.
   
\end{itemize}

\section{Background}

Let $p(\vz |\vx)$ be a Gaussian noise channel with $\vz = \csig \vx + \cnoise \eps$ and $\eps \sim \mathcal N(0, \mathbb I)$, where $\snr$ represents the Signal-to-Noise Ratio (SNR). 
The unknown data distribution is $p(\vx)$. 
The MMSE refers to the minimum mean square error for recovering $\vx$ in this noisy channel \citep{itd}.
\begin{equation}\label{eq:mmse} 
\mmse_x(\snr) \equiv \min_{\xhat(\vz, \snr)} \mathbb E_{p(\vz, \vx)} [ (\vx - \xhat(\vz, \snr))^2]
\end{equation}
The optimal estimator, $\xhat^*$, can be derived via variational calculus and written analytically.
\begin{equation}\label{eq:opt}
\xhat^*(\vz, \snr) \equiv \arg \mmse_x(\snr)  = \mathbb E_{\vx \sim p(\vx |\vz) }[ \vx]
\end{equation}
Sampling from the true posterior is typically intractable, but by using a neural network to solve Eq.~\ref{eq:mmse} we can get an approximation for $\xhat^*$. 
We also introduce pointwise MMSE which is just the MMSE evaluated at a single point $\vx$, and $\mathbb E_{p(\vx)}[\mmse(\vx |\snr)] = \mmse_x(\snr)$. 
\begin{equation}\label{eq:pmmse} 
\mmse(\vx | \snr) \equiv \mathbb E_{p(\vz|\vx)} [ (\vx - \xhat^*(\vz, \snr))^2] 
\end{equation}

From \cite{itd} we see that log likelihood can be written \emph{exactly} in terms of an expression that depends only on the MMSE solution to the Gaussian denoising problem. 
  \begin{empheq}[box=\fbox]{equation}\label{eq:density_simple} 
-\log p(\vx) = d/2 \log(2 \pi e) - \half \int_{0}^{\infty} d\snr \left(  \frac{d}{1+\snr} - \mmse(\vx | \snr) \right)
  \end{empheq}

We can use this result to derive elegant formulations for supervised learning and mutual information estimation. 

\section{Method}
\subsection{Mutual Information Estimation}
Consider that our data is drawn from some unknown joint distribution, $p(\vx, y)$. At this point we don't specify the domain of $y$, it could be discrete or continuous, vector or scalar. The pointwise denoising relation, Eq.~\ref{eq:density_simple}, holds for \emph{any} input distribution, so a valid choice would be $p(\vx|y)$. Therefore we can write a conditional version that holds for each $y$. 
  \begin{equation}\label{eq:cond_density} 
-\log p(\vx|y) = d/2 \log(2 \pi e) - \half \int_{0}^{\infty} d\snr \left(  \frac{d}{1+\snr} - \mmse(\vx | \snr, y) \right)
  \end{equation}
We use the following definitions for conditional MMSE.
\begin{align}\label{eq:cond_pmmse} 
\xhat^*(\vz, \snr, y) &\equiv \arg\min_{\xhat(\vz, \snr, y)} \mathbb E_{p(\vz|\vx) p(\vx,y)} [ (\vx - \xhat(\vz, \snr, y))^2] \nonumber\\
\mmse(\vx | \snr, y) &\equiv \mathbb E_{p(\vz|\vx)} [ (\vx - \xhat^*(\vz, \snr, y))^2]
\end{align}
We write the expected conditional MMSE as $\mmse_{x|y}(\snr) = \mathbb E_{p(\vx, y)} [\mmse(\vx | \snr, y)]$.

Now we can subtract Eq.~\ref{eq:density_simple} and Eq.~\ref{eq:cond_density} to get the following. 
  \begin{equation}\label{eq:point_mi} 
\log p(\vx|y) - \log p(\vx) =  \half \int_{0}^{\infty} d\snr \left(\mmse(\vx | \snr) - \mmse(\vx | \snr, y) \right)
  \end{equation}
The expression on the left is sometimes called \emph{pointwise mutual information}, because its expectation, $\mathbb E[\log p(\vx|y) / p(\vx)] = I(\vx ; y)$ is equal to the mutual information. 
  \begin{empheq}[box=\fbox]{equation}\label{eq:mi} 
I(\vx;y) =  \half \int_{0}^{\infty} d\snr \left(\mmse_x(\snr) - \mmse_{x|y}(\snr) \right)
  \end{empheq}
Conditioning on $y$ can only decrease the MMSE~\cite{wu2011functional}, so the mutual information is non-negative as expected. While Eq.~\ref{eq:point_mi} appears to be novel, a version of Eq.~\ref{eq:mi} appeared in \cite{guo}. 
This result holds for discrete or continuous data~\cite{guo}, or even mixed continuous and discrete, a particularly challenging problem~\cite{gao2017estimating}.
We propose to use recent advances in denoising diffusion modeling to approximate the right-hand side to achieve better mutual information estimates. 

\begin{figure}
	\begin{minipage}[t]{0.5\linewidth}
		\centering
		\includegraphics[width=\columnwidth]{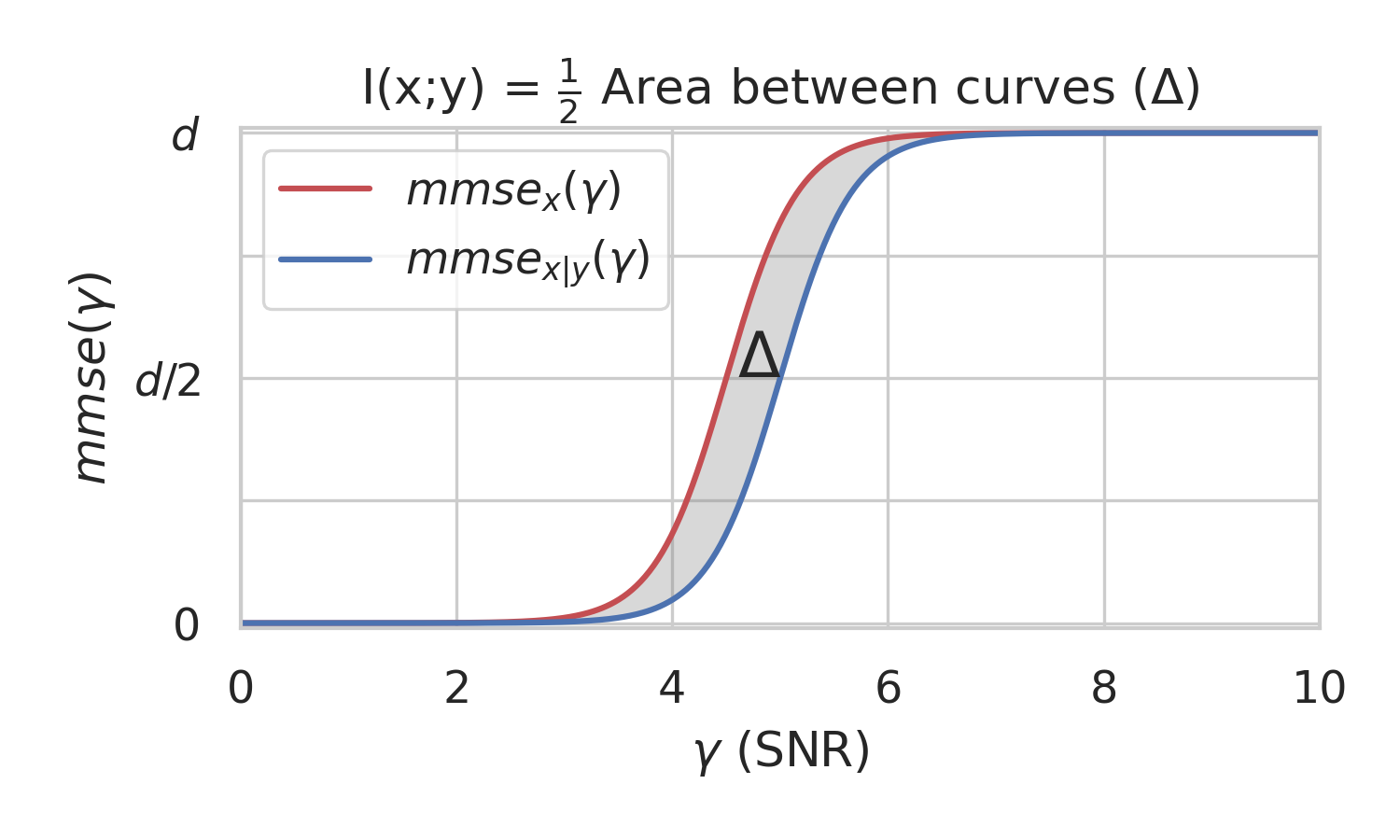}
		\caption{The mutual information is exactly half the area between MMSE curves for conditional and unconditional denoising. We use denoising diffusion models to approximate the MMSE curves, then numerically integrate to get an estimate of the mutual information. 
  }
		\label{fig:mind}
	\end{minipage}
	\begin{minipage}[t]{0.5\linewidth}
		\centering
		\includegraphics[width=\columnwidth]{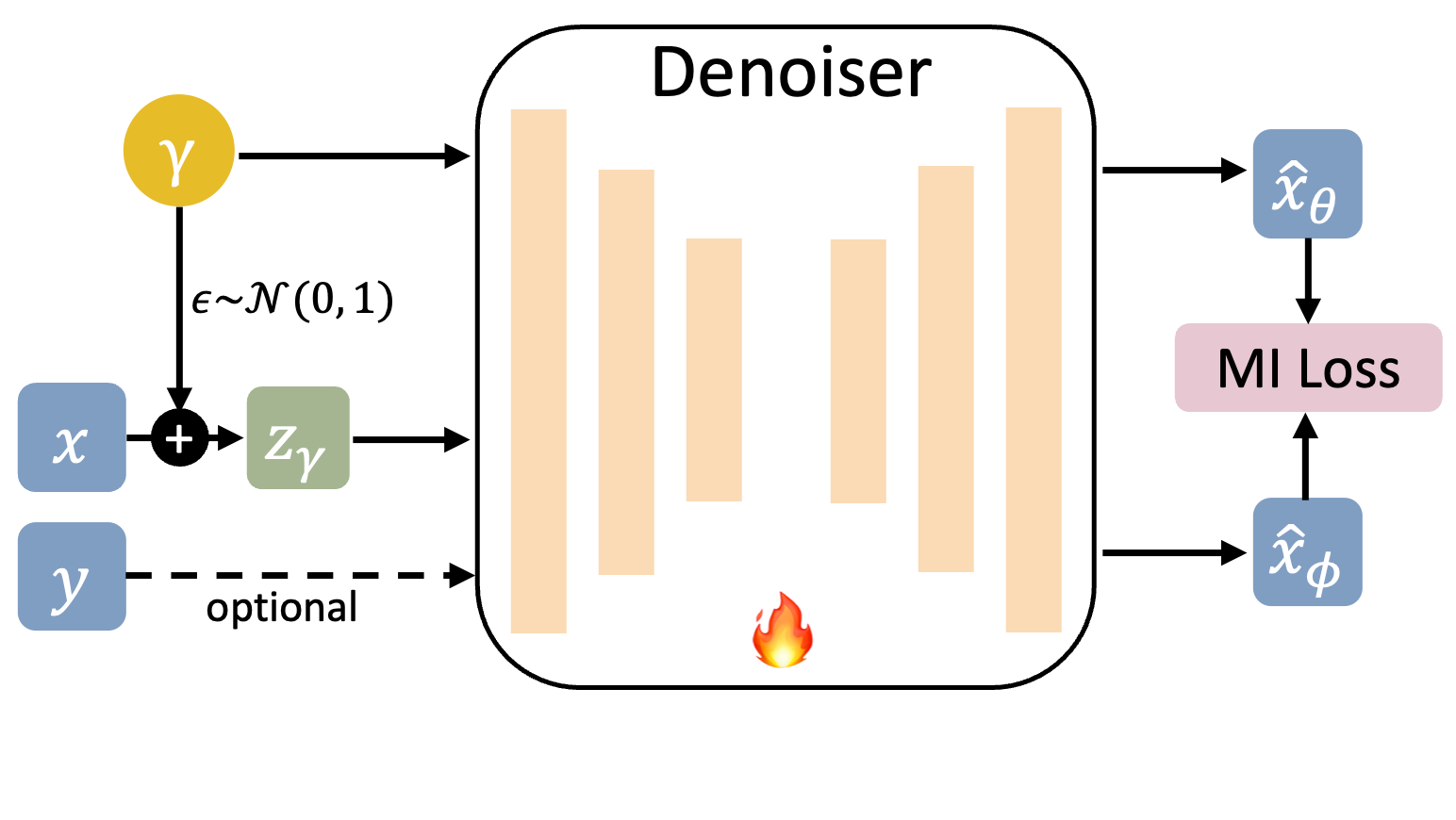}
		\caption{The schematic of the MMG training process. Noise at a given $\snr$ level is added to the data $\vx$, and a denoiser is used to recover it, with or without conditioning on $y$ at a 50\% probability. Finally, the MI loss, as defined in Eq.~\ref{eq:classifier_train}, is computed to backpropagate the gradient.}
		\label{fig:schematic}
	\end{minipage}
\end{figure}

To efficiently estimate the integral in Eq.~\ref{eq:mi}, we applied importance sampling in the integral on the right-hand side of Eq.~\ref{eq:mi} with the importance weights $q(\snr)$. 
\begin{empheq}[box=\fbox]{equation}\label{eq:mi-imp} 
    I(\vx;y) =  \half \mathbb E_{\snr, x, y}\left[ \left(\mmse_x(\snr) - \mmse_{x|y}(\snr) \right) / q(\snr)\right]
\end{empheq}
Therefore, we can train expert denoisers to estimate the MMSE for various distributions of $q(\snr)$, as different density consistently lead to distinct distributions of the MMSE gap in Figure~\ref{fig:mind}.

\subsection{Model Training}
In practice, we parametrize the denoising function in terms of a neural network. At training time, we have to solve two minimization problems, or actually a continuum of minimization problems for each SNR level. 
\begin{align}\label{eq:classifier_train}
\theta^* &= \arg \min_\theta \mathbb E_{p(\vz |\vx) p(\vx, y)} [ (\vx - \xhat_\theta(\vz, \snr, y))^2] \\
\phi^* &= \arg \min_\phi \mathbb E_{p(\vz |\vx) p(\vx)} [ (\vx - \xhat_\phi(\vz, \snr))^2]  \nonumber
\end{align}
%
%
We parametrize $\xhat_\theta(\vz, \snr, y)$ as some neural network that is trained with a mean square error loss to recover the signal, $\vx$, from noise and the auxiliary signal, $y$, across different SNRs, $\snr$ (see Figure~\ref{fig:schematic}). In principle we should train a separate neural network that recovers data from noise without conditioning on $y$. However, we can borrow from existing architectures which train a single network for both conditional and unconditional denoising~\cite{ho2021classifier}. Conditional diffusion models have shown compelling results in \cite{glide,dalle2,imagen}. During training we simply replace $y$ with a null value, $y=\emptyset$, some fraction of the time so that the same (conditional) network can learn to denoise in the absence of conditioning.  

In practice, we can't guarantee that the neural network finds the true global optimum, the MMSEs in Eq.~\ref{eq:mi}. Because MMSEs appear with both signs in that expression, we can't even guarantee that our network gives an upper or lower bound. 
While this is unfortunate, conventional wisdom suggests that sufficiently expressive neural networks do converge to global optima \citep{du2019gradient,jacot2018neural}.


\section{Implementation}\label{sec:implementation}
The core of the experiment is training a denoiser capable of effectively performing denoising across different noise levels, characterized by varying SNRs, to simulate an MMSE curve. The MI is then computed from the gap between the conditional and unconditional MMSE curves (Eq.~\ref{eq:mi}). This MMSE-based theoretical approach offers greater flexibility to enhance the accuracy and robustness of MI estimation. To leverage this flexibility for enhanced accuracy and robustness, we introduce the core components of our implementation below.


\subsection{Adaptive Importance Sampling}
As in \cite{itd}, we estimate integrals using importance weighted Monte Carlo estimators, with log-SNR values chosen from a logistic distribution with some location, $\mu$ and scale, $\sigma$. Since the shape of the integration area ($\Delta$ area in Fig. \ref{fig:mind}) is data-dependent, a fixed sampling distribution can be inefficient. For our adaptive variants, we therefore optimize this distribution for each specific task. 

To find the data-adapted parameters $(\mu, \sigma)$, we employ a two-stage procedure. First, we train a preliminary model and analyze its conditional MMSE curve, $\text{MMSE}_{x|y}$. Inspired by the use of error landmarks in density estimation \citep{itd}, our heuristic identifies the critical transition region of this curve, where the denoiser becomes effective. Specifically, we define the parameters as follows (where $d$ is the data dimensionality):
\begin{itemize}
    \item The location $\mu$ is set to the log-SNR where the MMSE curve crosses the $d/2$ error threshold. This centers our sampling distribution on the midpoint of the denoiser's transition from high to low error.
    \item The scale $\sigma$ is derived from the log-SNR where the curve crosses the $d/4$ threshold, capturing the steepness of this transition. For example, it can be set as the difference between the log-SNRs at the $d/2$ and $d/4$ crossings.
\end{itemize}
The final adaptive estimators are then trained with this optimized distribution, targeting the critical SNR range to yield more accurate and efficient MI estimates.

\subsection{The Orthogonal Principle}
To further enhance estimator stability, we incorporate the orthogonality principle from \cite{kong2023interpretable}. The principle states that for optimal denoisers, the gap in MMSE is precisely the expected squared distance between the conditional and unconditional estimators. This provides an alternative way to express the MI integrand. Formally, let $\hat{\vx}(\vz) = \mathbb{E}[\vx|\vz]$ and $\hat{\vx}(\vz, y) = \mathbb{E}[\vx|\vz, y]$. The identity at a fixed SNR is:
\begin{equation}\label{eq:ortho}
\underbrace{\mathbb{E}[\|\vx - \hat{\vx}(\vz)\|^2] - \mathbb{E}[\|\vx - \hat{\vx}(\vz, y)\|^2]}_{\text{MMSE Gap}} = \underbrace{\mathbb{E}[\|\hat{\vx}(\vz, y) - \hat{\vx}(\vz)\|^2]}_{\text{Orthogonal Form of the Gap}}
\end{equation}
After expanding the left-hand side, the cross terms vanish due to the orthogonality principle \cite{kong2023interpretable}, resulting in the expression on the right-hand side. This allows us to substitute the original integrand in our MI formula (Eq.~\ref{eq:mi}) with the term on the right-hand side. The orthogonal MI estimator, in its practical importance-sampled form, is therefore:
\begin{equation}\label{eq:mi_ortho}
I(\vx;y) = \frac{1}{2} \mathbb{E}_{\vx, y, \vz, \snr \sim q} \left[ \frac{\|\hat{\vx}(\vz, y) - \hat{\vx}(\vz)\|^2}{q(\snr)} \right]
\end{equation}
In practice, this is implemented as a \textit{training-free, inference-time plug-in}. This formulation guarantees a non-negative integrand and improves the stability of MI estimates, as detailed in Appendix~\ref{sec:ortho_proof}.

The practical advantage of this formulation is its numerical stability. The standard MMSE gap is computed as a difference between two large, separately estimated MSE values. Small approximation errors in each term can lead to a noisy, high-variance result for their difference, which can even become negative. In contrast, the orthogonal form computes the integrand as a single, squared term, which is guaranteed to be non-negative and is empirically much smoother. 

\section{Experiments} 
We evaluate four variants of our estimator in the experiments for ablation comparison:
\begin{itemize}[left=0pt]
    \item \textbf{MMG}: The baseline estimator using a fixed, default importance sampling distribution.
    \item \textbf{MMG-adaptive}: Employs adaptive importance sampling for more accurate integration.
    \item \textbf{MMG-orthogonal}: Applies the orthogonal principle at inference time with baseline sampling.
    \item \textbf{MMG-orthogonal-adaptive}: Combines both adaptive sampling and the orthogonal principle.
\end{itemize}
\subsection{MI Estimation Benchmark}
We evaluate on the benchmark of \citep{czyż2023normalevaluationmutualinformation}, which combines base distributions (Uniform, Normal with dense or sparse correlation, and long-tailed Student-t) with MI-preserving nonlinear transformations (Half-Cube, Asinh, Swiss-roll, Spiral). This setup introduces high dimensionality, heavy tails, sparsity, and non-linear geometry. We compare MMG against neural estimators, such as \textsc{mine}~\citep{belghazi2018mine}, 
\textsc{Infonce}~\citep{oord2019representation},\textsc{nwj}~\citep{nguyen2007neurips}, \textsc{doe}~\citep{mcallester2020formal} and \textsc{MINDE} \cite{franzese2023minde}. 

The results in Table~\ref{tab:benchmark} establish our method's state-of-the-art performance. Our \textbf{MMG-orthogonal-adaptive} and \textbf{MMG-orthogonal} variants succeed on 39/40 and 37/40 tasks respectively, surpassing the MINDE (35/40). This robustness is rooted in our two main contributions: adaptive importance sampling ensures accuracy by focusing the integral on critical SNR regions, while the orthogonal principle guarantees a low-variance integrand for stability. This combination allows our method to excel on complex non-linear datasets where traditional methods fail.

\renewcommand{\tabcolsep}{0.0pt}
\begin{table}[H]
\vspace*{-0cm}
\tiny

\resizebox{\textwidth}{!}{
\begin{tabular}{lrrrrrrrrrrrrrrrrrrrrrrrrrrrrrrrrrrrrrrrr}
\toprule

GT & 0.2 & 0.4 & 0.3 & 0.4 & 0.4 & 0.4 & 0.4 & 1.0 & 1.0 & 1.0 & 1.0 & 0.3 & 1.0 & 1.3 & 1.0 & 0.4 & 1.0 & 0.6 & 1.6 & 0.4 & 1.0 & 1.0 & 1.0 & 1.0 & 1.0 & 1.0 & 1.0 & 1.0 & 1.0 & 0.2 & 0.4 & 0.2 & 0.3 & 0.2 & 0.4 & 0.3 & 0.4 & 1.7 & 0.3 & 0.4 \\
\midrule

MINE & {\cellcolor[HTML]{F2F2F2}} 0.2 & {\cellcolor[HTML]{F2F2F2}} 0.4 & {\cellcolor[HTML]{EED3DB}} 0.2 & {\cellcolor[HTML]{F2F2F2}} 0.4 & {\cellcolor[HTML]{F2F2F2}} 0.4 & {\cellcolor[HTML]{F2F2F2}} 0.4 & {\cellcolor[HTML]{F2F2F2}} 0.4 & {\cellcolor[HTML]{F2F2F2}} 1.0 & {\cellcolor[HTML]{F2F2F2}} 1.0 & {\cellcolor[HTML]{F2F2F2}} 1.0 & {\cellcolor[HTML]{F2F2F2}} 1.0 & {\cellcolor[HTML]{F2F2F2}} 0.3 & {\cellcolor[HTML]{F2F2F2}} 1.0 & {\cellcolor[HTML]{F2F2F2}} 1.3 & {\cellcolor[HTML]{F2F2F2}} 1.0 & {\cellcolor[HTML]{F2F2F2}} 0.4 & {\cellcolor[HTML]{F2F2F2}} 1.0 & {\cellcolor[HTML]{F2F2F2}} 0.6 & {\cellcolor[HTML]{F2F2F2}} 1.6 & {\cellcolor[HTML]{F2F2F2}} 0.4 & {\cellcolor[HTML]{EED3DB}} 0.9 & {\cellcolor[HTML]{EED3DB}} 0.9 & {\cellcolor[HTML]{EED3DB}} 0.9 & {\cellcolor[HTML]{E7ADBE}} 0.8 & {\cellcolor[HTML]{E188A2}} 0.7 & {\cellcolor[HTML]{DB6185}} 0.6 & {\cellcolor[HTML]{EED3DB}} 0.9 & {\cellcolor[HTML]{EED3DB}} 0.9 & {\cellcolor[HTML]{EED3DB}} 0.9 & {\cellcolor[HTML]{E7ADBE}} 0.0 & {\cellcolor[HTML]{DB6185}} 0.0 & {\cellcolor[HTML]{EED3DB}} 0.1 & {\cellcolor[HTML]{E7ADBE}} 0.1 & {\cellcolor[HTML]{EED3DB}} 0.1 & {\cellcolor[HTML]{E7ADBE}} 0.2 & {\cellcolor[HTML]{EED3DB}} 0.2 & {\cellcolor[HTML]{F2F2F2}} 0.4 & {\cellcolor[HTML]{F2F2F2}} 1.7 & {\cellcolor[HTML]{F2F2F2}} 0.3 & {\cellcolor[HTML]{F2F2F2}} 0.4 \\
InfoNCE & {\cellcolor[HTML]{F2F2F2}} 0.2 & {\cellcolor[HTML]{F2F2F2}} 0.4 & {\cellcolor[HTML]{F2F2F2}} 0.3 & {\cellcolor[HTML]{F2F2F2}} 0.4 & {\cellcolor[HTML]{F2F2F2}} 0.4 & {\cellcolor[HTML]{F2F2F2}} 0.4 & {\cellcolor[HTML]{F2F2F2}} 0.4 & {\cellcolor[HTML]{F2F2F2}} 1.0 & {\cellcolor[HTML]{F2F2F2}} 1.0 & {\cellcolor[HTML]{F2F2F2}} 1.0 & {\cellcolor[HTML]{F2F2F2}} 1.0 & {\cellcolor[HTML]{F2F2F2}} 0.3 & {\cellcolor[HTML]{F2F2F2}} 1.0 & {\cellcolor[HTML]{F2F2F2}} 1.3 & {\cellcolor[HTML]{F2F2F2}} 1.0 & {\cellcolor[HTML]{F2F2F2}} 0.4 & {\cellcolor[HTML]{F2F2F2}} 1.0 & {\cellcolor[HTML]{F2F2F2}} 0.6 & {\cellcolor[HTML]{F2F2F2}} 1.6 & {\cellcolor[HTML]{F2F2F2}} 0.4 & {\cellcolor[HTML]{EED3DB}} 0.9 & {\cellcolor[HTML]{F2F2F2}} 1.0 & {\cellcolor[HTML]{F2F2F2}} 1.0 & {\cellcolor[HTML]{E7ADBE}} 0.8 & {\cellcolor[HTML]{E7ADBE}} 0.8 & {\cellcolor[HTML]{E7ADBE}} 0.8 & {\cellcolor[HTML]{EED3DB}} 0.9 & {\cellcolor[HTML]{F2F2F2}} 1.0 & {\cellcolor[HTML]{F2F2F2}} 1.0 & {\cellcolor[HTML]{F2F2F2}} 0.2 & {\cellcolor[HTML]{EED3DB}} 0.3 & {\cellcolor[HTML]{F2F2F2}} 0.2 & {\cellcolor[HTML]{F2F2F2}} 0.3 & {\cellcolor[HTML]{F2F2F2}} 0.2 & {\cellcolor[HTML]{F2F2F2}} 0.4 & {\cellcolor[HTML]{F2F2F2}} 0.3 & {\cellcolor[HTML]{F2F2F2}} 0.4 & {\cellcolor[HTML]{F2F2F2}} 1.7 & {\cellcolor[HTML]{F2F2F2}} 0.3 & {\cellcolor[HTML]{F2F2F2}} 0.4 \\
D-V & {\cellcolor[HTML]{F2F2F2}} 0.2 & {\cellcolor[HTML]{F2F2F2}} 0.4 & {\cellcolor[HTML]{F2F2F2}} 0.3 & {\cellcolor[HTML]{F2F2F2}} 0.4 & {\cellcolor[HTML]{F2F2F2}} 0.4 & {\cellcolor[HTML]{F2F2F2}} 0.4 & {\cellcolor[HTML]{F2F2F2}} 0.4 & {\cellcolor[HTML]{F2F2F2}} 1.0 & {\cellcolor[HTML]{F2F2F2}} 1.0 & {\cellcolor[HTML]{F2F2F2}} 1.0 & {\cellcolor[HTML]{F2F2F2}} 1.0 & {\cellcolor[HTML]{F2F2F2}} 0.3 & {\cellcolor[HTML]{F2F2F2}} 1.0 & {\cellcolor[HTML]{F2F2F2}} 1.3 & {\cellcolor[HTML]{F2F2F2}} 1.0 & {\cellcolor[HTML]{F2F2F2}} 0.4 & {\cellcolor[HTML]{F2F2F2}} 1.0 & {\cellcolor[HTML]{F2F2F2}} 0.6 & {\cellcolor[HTML]{F2F2F2}} 1.6 & {\cellcolor[HTML]{F2F2F2}} 0.4 & {\cellcolor[HTML]{EED3DB}} 0.9 & {\cellcolor[HTML]{F2F2F2}} 1.0 & {\cellcolor[HTML]{F2F2F2}} 1.0 & {\cellcolor[HTML]{E7ADBE}} 0.8 & {\cellcolor[HTML]{E7ADBE}} 0.8 & {\cellcolor[HTML]{E7ADBE}} 0.8 & {\cellcolor[HTML]{EED3DB}} 0.9 & {\cellcolor[HTML]{F2F2F2}} 1.0 & {\cellcolor[HTML]{F2F2F2}} 1.0 & {\cellcolor[HTML]{E7ADBE}} 0.0 & {\cellcolor[HTML]{DB6185}} 0.0 & {\cellcolor[HTML]{EED3DB}} 0.1 & {\cellcolor[HTML]{E7ADBE}} 0.1 & {\cellcolor[HTML]{F2F2F2}} 0.2 & {\cellcolor[HTML]{E7ADBE}} 0.2 & {\cellcolor[HTML]{EED3DB}} 0.2 & {\cellcolor[HTML]{F2F2F2}} 0.4 & {\cellcolor[HTML]{F2F2F2}} 1.7 & {\cellcolor[HTML]{F2F2F2}} 0.3 & {\cellcolor[HTML]{F2F2F2}} 0.4 \\
NWJ & {\cellcolor[HTML]{F2F2F2}} 0.2 & {\cellcolor[HTML]{F2F2F2}} 0.4 & {\cellcolor[HTML]{F2F2F2}} 0.3 & {\cellcolor[HTML]{F2F2F2}} 0.4 & {\cellcolor[HTML]{F2F2F2}} 0.4 & {\cellcolor[HTML]{F2F2F2}} 0.4 & {\cellcolor[HTML]{F2F2F2}} 0.4 & {\cellcolor[HTML]{F2F2F2}} 1.0 & {\cellcolor[HTML]{F2F2F2}} 1.0 & {\cellcolor[HTML]{F2F2F2}} 1.0 & {\cellcolor[HTML]{F2F2F2}} 1.0 & {\cellcolor[HTML]{F2F2F2}} 0.3 & {\cellcolor[HTML]{F2F2F2}} 1.0 & {\cellcolor[HTML]{F2F2F2}} 1.3 & {\cellcolor[HTML]{F2F2F2}} 1.0 & {\cellcolor[HTML]{F2F2F2}} 0.4 & {\cellcolor[HTML]{F2F2F2}} 1.0 & {\cellcolor[HTML]{F2F2F2}} 0.6 & {\cellcolor[HTML]{F2F2F2}} 1.6 & {\cellcolor[HTML]{F2F2F2}} 0.4 & {\cellcolor[HTML]{EED3DB}} 0.9 & {\cellcolor[HTML]{F2F2F2}} 1.0 & {\cellcolor[HTML]{F2F2F2}} 1.0 & {\cellcolor[HTML]{E7ADBE}} 0.8 & {\cellcolor[HTML]{E7ADBE}} 0.8 & {\cellcolor[HTML]{E7ADBE}} 0.8 & {\cellcolor[HTML]{EED3DB}} 0.9 & {\cellcolor[HTML]{F2F2F2}} 1.0 & {\cellcolor[HTML]{F2F2F2}} 1.0 & {\cellcolor[HTML]{E7ADBE}} 0.0 & {\cellcolor[HTML]{DB6185}} 0.0 & {\cellcolor[HTML]{E7ADBE}} 0.0 & {\cellcolor[HTML]{D53D69}} -0.6 & {\cellcolor[HTML]{EED3DB}} 0.1 & {\cellcolor[HTML]{E188A2}} 0.1 & {\cellcolor[HTML]{EED3DB}} 0.2 & {\cellcolor[HTML]{F2F2F2}} 0.4 & {\cellcolor[HTML]{F2F2F2}} 1.7 & {\cellcolor[HTML]{F2F2F2}} 0.3 & {\cellcolor[HTML]{F2F2F2}} 0.4 \\

DoE(Gaussian) & {\cellcolor[HTML]{F2F2F2}} 0.2 & {\cellcolor[HTML]{DAE1EB}} 0.5 & {\cellcolor[HTML]{F2F2F2}} 0.3 & {\cellcolor[HTML]{BCCCE1}} 0.6 & {\cellcolor[HTML]{F2F2F2}} 0.4 & {\cellcolor[HTML]{F2F2F2}} 0.4 & {\cellcolor[HTML]{F2F2F2}} 0.4 & {\cellcolor[HTML]{E49AAF}} 0.7 & {\cellcolor[HTML]{F2F2F2}} 1.0 & {\cellcolor[HTML]{F2F2F2}} 1.0 & {\cellcolor[HTML]{F2F2F2}} 1.0 & {\cellcolor[HTML]{DAE1EB}} 0.4 & {\cellcolor[HTML]{E49AAF}} 0.7 & {\cellcolor[HTML]{4479BB}} 7.8 & {\cellcolor[HTML]{F2F2F2}} 1.0 & {\cellcolor[HTML]{BCCCE1}} 0.6 & {\cellcolor[HTML]{EED9DF}} 0.9 & {\cellcolor[HTML]{4479BB}} 1.3 &  & {\cellcolor[HTML]{F2F2F2}} 0.4 & {\cellcolor[HTML]{E49AAF}} 0.7 & {\cellcolor[HTML]{F2F2F2}} 1.0 & {\cellcolor[HTML]{F2F2F2}} 1.0 & {\cellcolor[HTML]{DA5C81}} 0.5 & {\cellcolor[HTML]{DF7B98}} 0.6 & {\cellcolor[HTML]{DF7B98}} 0.6 & {\cellcolor[HTML]{DF7B98}} 0.6 & {\cellcolor[HTML]{E49AAF}} 0.7 & {\cellcolor[HTML]{E9BAC8}} 0.8 & {\cellcolor[HTML]{4479BB}} 6.7 & {\cellcolor[HTML]{4479BB}} 7.9 & {\cellcolor[HTML]{4479BB}} 1.8 & {\cellcolor[HTML]{4479BB}} 2.5 & {\cellcolor[HTML]{7FA2CE}} 0.6 & {\cellcolor[HTML]{4479BB}} 4.2 & {\cellcolor[HTML]{4479BB}} 1.2 & {\cellcolor[HTML]{EED9DF}} 1.6 & {\cellcolor[HTML]{E9BAC8}} 0.1 & {\cellcolor[HTML]{F2F2F2}} 0.4 &  \\
DoE(Logistic) & {\cellcolor[HTML]{EED9DF}} 0.1 & {\cellcolor[HTML]{F2F2F2}} 0.4 & {\cellcolor[HTML]{EED9DF}} 0.2 & {\cellcolor[HTML]{F2F2F2}} 0.4 & {\cellcolor[HTML]{F2F2F2}} 0.4 & {\cellcolor[HTML]{F2F2F2}} 0.4 & {\cellcolor[HTML]{F2F2F2}} 0.4 & {\cellcolor[HTML]{DF7B98}} 0.6 & {\cellcolor[HTML]{EED9DF}} 0.9 & {\cellcolor[HTML]{EED9DF}} 0.9 & {\cellcolor[HTML]{F2F2F2}} 1.0 & {\cellcolor[HTML]{F2F2F2}} 0.3 & {\cellcolor[HTML]{E49AAF}} 0.7 & {\cellcolor[HTML]{4479BB}} 7.8 & {\cellcolor[HTML]{F2F2F2}} 1.0 & {\cellcolor[HTML]{BCCCE1}} 0.6 & {\cellcolor[HTML]{EED9DF}} 0.9 & {\cellcolor[HTML]{4479BB}} 1.3 &  & {\cellcolor[HTML]{F2F2F2}} 0.4 & {\cellcolor[HTML]{E9BAC8}} 0.8 & {\cellcolor[HTML]{DAE1EB}} 1.1 & {\cellcolor[HTML]{F2F2F2}} 1.0 & {\cellcolor[HTML]{DA5C81}} 0.5 & {\cellcolor[HTML]{DF7B98}} 0.6 & {\cellcolor[HTML]{DF7B98}} 0.6 & {\cellcolor[HTML]{E49AAF}} 0.7 & {\cellcolor[HTML]{E9BAC8}} 0.8 & {\cellcolor[HTML]{E9BAC8}} 0.8 &  & {\cellcolor[HTML]{4479BB}} 2.0 & {\cellcolor[HTML]{9DB7D7}} 0.5 & {\cellcolor[HTML]{628DC4}} 0.8 & {\cellcolor[HTML]{DAE1EB}} 0.3 & {\cellcolor[HTML]{4479BB}} 1.5 & {\cellcolor[HTML]{9DB7D7}} 0.6 & {\cellcolor[HTML]{EED9DF}} 1.6 & {\cellcolor[HTML]{E9BAC8}} 0.1 & {\cellcolor[HTML]{F2F2F2}} 0.4 & \\
\midrule

\textbf{MINDE--\textsc{j} ($\sigma=1$)} \ & {\cellcolor[HTML]{F2F2F2}} 0.2 & {\cellcolor[HTML]{F2F2F2}} 0.4 & {\cellcolor[HTML]{F2F2F2}} 0.3 & {\cellcolor[HTML]{F2F2F2}} 0.4 & {\cellcolor[HTML]{F2F2F2}} 0.4 & {\cellcolor[HTML]{F2F2F2}} 0.4 & {\cellcolor[HTML]{F2F2F2}} 0.4 & {\cellcolor[HTML]{D4DDE9}} 1.1 & {\cellcolor[HTML]{F2F2F2}} 1.0 & {\cellcolor[HTML]{F2F2F2}} 1.0 & {\cellcolor[HTML]{F2F2F2}} 1.0 & {\cellcolor[HTML]{F2F2F2}} 0.3 & {\cellcolor[HTML]{EED3DB}} 0.9 & {\cellcolor[HTML]{EED3DB}} 1.2 & {\cellcolor[HTML]{F2F2F2}} 1.0 & {\cellcolor[HTML]{F2F2F2}} 0.4 & {\cellcolor[HTML]{F2F2F2}} 1.0 & {\cellcolor[HTML]{F2F2F2}} 0.6 & {\cellcolor[HTML]{D4DDE9}} 1.7 & {\cellcolor[HTML]{F2F2F2}} 0.4 & {\cellcolor[HTML]{F2F2F2}} 1.0 & {\cellcolor[HTML]{F2F2F2}} 1.0 & {\cellcolor[HTML]{F2F2F2}} 1.0 & {\cellcolor[HTML]{EED3DB}} 0.9 & {\cellcolor[HTML]{EED3DB}} 0.9 & {\cellcolor[HTML]{EED3DB}} 0.9 & {\cellcolor[HTML]{F2F2F2}} 1.0 & {\cellcolor[HTML]{EED3DB}} 0.9 & {\cellcolor[HTML]{F2F2F2}} 1.0 & {\cellcolor[HTML]{F2F2F2}} 0.2 & {\cellcolor[HTML]{F2F2F2}} 0.4 & {\cellcolor[HTML]{F2F2F2}} 0.2 & {\cellcolor[HTML]{F2F2F2}} 0.3 & {\cellcolor[HTML]{F2F2F2}} 0.2 & {\cellcolor[HTML]{D4DDE9}} 0.5 & {\cellcolor[HTML]{F2F2F2}} 0.3 & {\cellcolor[HTML]{D4DDE9}} 0.5 & {\cellcolor[HTML]{EED3DB}} 1.6 & {\cellcolor[HTML]{F2F2F2}} 0.3 & {\cellcolor[HTML]{F2F2F2}} 0.4 \\

\textbf{MINDE--\textsc{j}}  & {\cellcolor[HTML]{F2F2F2}} 0.2 & {\cellcolor[HTML]{F2F2F2}} 0.4 & {\cellcolor[HTML]{F2F2F2}} 0.3 & {\cellcolor[HTML]{F2F2F2}} 0.4 & {\cellcolor[HTML]{F2F2F2}} 0.4 & {\cellcolor[HTML]{F2F2F2}} 0.4 & {\cellcolor[HTML]{F2F2F2}} 0.4 & {\cellcolor[HTML]{AFC4DD}} 1.2 & {\cellcolor[HTML]{F2F2F2}} 1.0 & {\cellcolor[HTML]{F2F2F2}} 1.0 & {\cellcolor[HTML]{F2F2F2}} 1.0 & {\cellcolor[HTML]{F2F2F2}} 0.3 & {\cellcolor[HTML]{F2F2F2}} 1.0 & {\cellcolor[HTML]{F2F2F2}} 1.3 & {\cellcolor[HTML]{F2F2F2}} 1.0 & {\cellcolor[HTML]{F2F2F2}} 0.4 & {\cellcolor[HTML]{F2F2F2}} 1.0 & {\cellcolor[HTML]{F2F2F2}} 0.6 & {\cellcolor[HTML]{D4DDE9}} 1.7 & {\cellcolor[HTML]{F2F2F2}} 0.4 & {\cellcolor[HTML]{D4DDE9}} 1.1 & {\cellcolor[HTML]{F2F2F2}} 1.0 & {\cellcolor[HTML]{F2F2F2}} 1.0 & {\cellcolor[HTML]{F2F2F2}} 1.0 & {\cellcolor[HTML]{EED3DB}} 0.9 & {\cellcolor[HTML]{EED3DB}} 0.9 & {\cellcolor[HTML]{D4DDE9}} 1.1 & {\cellcolor[HTML]{F2F2F2}} 1.0 & {\cellcolor[HTML]{F2F2F2}} 1.0 & {\cellcolor[HTML]{EED3DB}} 0.1 & {\cellcolor[HTML]{E7ADBE}} 0.2 & {\cellcolor[HTML]{F2F2F2}} 0.2 & {\cellcolor[HTML]{F2F2F2}} 0.3 & {\cellcolor[HTML]{F2F2F2}} 0.2 & {\cellcolor[HTML]{D4DDE9}} 0.5 & {\cellcolor[HTML]{F2F2F2}} 0.3 & {\cellcolor[HTML]{F2F2F2}} 0.4 & {\cellcolor[HTML]{F2F2F2}} 1.7 & {\cellcolor[HTML]{F2F2F2}} 0.3 & {\cellcolor[HTML]{F2F2F2}} 0.4 \\

\textbf{MINDE--\textsc{c} ($\sigma=1$)} & {\cellcolor[HTML]{F2F2F2}} 0.2 & {\cellcolor[HTML]{F2F2F2}} 0.4 & {\cellcolor[HTML]{F2F2F2}} 0.3 & {\cellcolor[HTML]{F2F2F2}} 0.4 & {\cellcolor[HTML]{F2F2F2}} 0.4 & {\cellcolor[HTML]{F2F2F2}} 0.4 & {\cellcolor[HTML]{F2F2F2}} 0.4 & {\cellcolor[HTML]{F2F2F2}} 1.0 & {\cellcolor[HTML]{F2F2F2}} 1.0 & {\cellcolor[HTML]{F2F2F2}} 1.0 & {\cellcolor[HTML]{F2F2F2}} 1.0 & {\cellcolor[HTML]{F2F2F2}} 0.3 & {\cellcolor[HTML]{F2F2F2}} 1.0 & {\cellcolor[HTML]{F2F2F2}} 1.3 & {\cellcolor[HTML]{F2F2F2}} 1.0 & {\cellcolor[HTML]{F2F2F2}} 0.4 & {\cellcolor[HTML]{F2F2F2}} 1.0 & {\cellcolor[HTML]{F2F2F2}} 0.6 & {\cellcolor[HTML]{F2F2F2}} 1.6 & {\cellcolor[HTML]{F2F2F2}} 0.4 & {\cellcolor[HTML]{EED9DF}} 0.9 & {\cellcolor[HTML]{F2F2F2}} 1.0 & {\cellcolor[HTML]{F2F2F2}} 1.0 & {\cellcolor[HTML]{EED9DF}} 0.9 & {\cellcolor[HTML]{EED9DF}} 0.9 & {\cellcolor[HTML]{EED9DF}} 0.9 & {\cellcolor[HTML]{EED9DF}} 0.9 & {\cellcolor[HTML]{F2F2F2}} 1.0 & {\cellcolor[HTML]{EED9DF}} 0.9 & {\cellcolor[HTML]{EED9DF}} 0.1 & {\cellcolor[HTML]{EED9DF}} 0.3 & {\cellcolor[HTML]{F2F2F2}} 0.2 & {\cellcolor[HTML]{F2F2F2}} 0.3 & {\cellcolor[HTML]{F2F2F2}} 0.2 & {\cellcolor[HTML]{F2F2F2}} 0.4 & {\cellcolor[HTML]{F2F2F2}} 0.3 & {\cellcolor[HTML]{EED9DF}} 0.3 & {\cellcolor[HTML]{F2F2F2}} 1.7 & {\cellcolor[HTML]{F2F2F2}} 0.3 & {\cellcolor[HTML]{F2F2F2}} 0.4 \\

\textbf{MINDE--\textsc{c}} & {\cellcolor[HTML]{F2F2F2}} 0.2 & {\cellcolor[HTML]{F2F2F2}} 0.4 & {\cellcolor[HTML]{F2F2F2}} 0.3 & {\cellcolor[HTML]{F2F2F2}} 0.4 & {\cellcolor[HTML]{F2F2F2}} 0.4 & {\cellcolor[HTML]{F2F2F2}} 0.4 & {\cellcolor[HTML]{F2F2F2}} 0.4 & {\cellcolor[HTML]{F2F2F2}} 1.0 & {\cellcolor[HTML]{F2F2F2}} 1.0 & {\cellcolor[HTML]{F2F2F2}} 1.0 & {\cellcolor[HTML]{F2F2F2}} 1.0 & {\cellcolor[HTML]{F2F2F2}} 0.3 & {\cellcolor[HTML]{F2F2F2}} 1.0 & {\cellcolor[HTML]{F2F2F2}} 1.3 & {\cellcolor[HTML]{F2F2F2}} 1.0 & {\cellcolor[HTML]{F2F2F2}} 0.4 & {\cellcolor[HTML]{F2F2F2}} 1.0 & {\cellcolor[HTML]{F2F2F2}} 0.6 & {\cellcolor[HTML]{F2F2F2}} 1.6 & {\cellcolor[HTML]{F2F2F2}} 0.4 & {\cellcolor[HTML]{F2F2F2}} 1.0 & {\cellcolor[HTML]{F2F2F2}} 1.0 & {\cellcolor[HTML]{F2F2F2}} 1.0 & {\cellcolor[HTML]{EED9DF}} 0.9 & {\cellcolor[HTML]{EED9DF}} 0.9 & {\cellcolor[HTML]{EED9DF}} 0.9 & {\cellcolor[HTML]{F2F2F2}} 1.0 & {\cellcolor[HTML]{F2F2F2}} 1.0 & {\cellcolor[HTML]{F2F2F2}} 1.0 & {\cellcolor[HTML]{EED9DF}} 0.1 & {\cellcolor[HTML]{EED9DF}} 0.3 & {\cellcolor[HTML]{F2F2F2}} 0.2 & {\cellcolor[HTML]{F2F2F2}} 0.3 & {\cellcolor[HTML]{F2F2F2}} 0.2 & {\cellcolor[HTML]{F2F2F2}} 0.4 & {\cellcolor[HTML]{F2F2F2}} 0.3 & {\cellcolor[HTML]{F2F2F2}} 0.4 & {\cellcolor[HTML]{F2F2F2}} 1.7 & {\cellcolor[HTML]{F2F2F2}} 0.3 & {\cellcolor[HTML]{F2F2F2}} 0.4 \\

\midrule

\textbf{MMG} & {\cellcolor[HTML]{F2F2F2}} 0.2 & {\cellcolor[HTML]{F2F2F2}} 0.4 & {\cellcolor[HTML]{F2F2F2}} 0.3 & {\cellcolor[HTML]{D4DDE9}} 0.5 & {\cellcolor[HTML]{F2F2F2}} 0.4 & {\cellcolor[HTML]{F2F2F2}} 0.4 & {\cellcolor[HTML]{F2F2F2}} 0.4 & {\cellcolor[HTML]{EED3DB}} 0.9 & {\cellcolor[HTML]{D4DDE9}} 1.1 & {\cellcolor[HTML]{EED3DB}} 0.9 & {\cellcolor[HTML]{F2F2F2}} 1.0 & {\cellcolor[HTML]{E7ADBE}} 0.2 & {\cellcolor[HTML]{EED3DB}} 0.9 & {\cellcolor[HTML]{EED3DB}} 1.2 & {\cellcolor[HTML]{F2F2F2}} 1.0 & {\cellcolor[HTML]{F2F2F2}} 0.4 & {\cellcolor[HTML]{F2F2F2}} 1.0 & {\cellcolor[HTML]{F2F2F2}} 0.6 & {\cellcolor[HTML]{E7ADBE}} 1.4 & {\cellcolor[HTML]{D4DDE9}} 0.5 & {\cellcolor[HTML]{F2F2F2}} 1.0 & {\cellcolor[HTML]{F2F2F2}} 1.0 & {\cellcolor[HTML]{F2F2F2}} 1.0 & {\cellcolor[HTML]{D4DDE9}} 1.1 & {\cellcolor[HTML]{F2F2F2}} 1.0 & {\cellcolor[HTML]{F2F2F2}} 1.0 & {\cellcolor[HTML]{D4DDE9}} 1.2 & {\cellcolor[HTML]{F2F2F2}} 1.0 & {\cellcolor[HTML]{F2F2F2}} 1.0 & {\cellcolor[HTML]{F2F2F2}} 0.2 & {\cellcolor[HTML]{E7ADBE}} 0.3 & {\cellcolor[HTML]{F2F2F2}} 0.2 & {\cellcolor[HTML]{F2F2F2}} 0.3 & {\cellcolor[HTML]{F2F2F2}} 0.2 & {\cellcolor[HTML]{F2F2F2}} 0.4 & {\cellcolor[HTML]{E7ADBE}} 0.2 & {\cellcolor[HTML]{E7ADBE}} 0.3 & {\cellcolor[HTML]{F2F2F2}} 1.7 & {\cellcolor[HTML]{F2F2F2}} 0.3 & {\cellcolor[HTML]{F2F2F2}} 0.4 \\

\textbf{MMG-adaptive} & {\cellcolor[HTML]{F2F2F2}} 0.2 & {\cellcolor[HTML]{F2F2F2}} 0.4 & {\cellcolor[HTML]{F2F2F2}} 0.3 & {\cellcolor[HTML]{F2F2F2}} 0.4 & {\cellcolor[HTML]{F2F2F2}} 0.4 & {\cellcolor[HTML]{F2F2F2}} 0.4 & {\cellcolor[HTML]{F2F2F2}} 0.4 & {\cellcolor[HTML]{EED3DB}} 0.9 & {\cellcolor[HTML]{F2F2F2}} 1.0 & {\cellcolor[HTML]{EED3DB}} 0.9 & {\cellcolor[HTML]{F2F2F2}} 1.0 & {\cellcolor[HTML]{F2F2F2}} 0.3 & {\cellcolor[HTML]{F2F2F2}} 1.0 & {\cellcolor[HTML]{EED3DB}} 1.2 & {\cellcolor[HTML]{F2F2F2}} 1.0 & {\cellcolor[HTML]{F2F2F2}} 0.4 & {\cellcolor[HTML]{F2F2F2}} 1.0 & {\cellcolor[HTML]{F2F2F2}} 0.6 & {\cellcolor[HTML]{E7ADBE}} 1.4 & {\cellcolor[HTML]{F2F2F2}} 0.4 & {\cellcolor[HTML]{F2F2F2}} 1.0 & {\cellcolor[HTML]{F2F2F2}} 1.0 & {\cellcolor[HTML]{D4DDE9}} 1.1 & {\cellcolor[HTML]{F2F2F2}} 1.0 & {\cellcolor[HTML]{F2F2F2}} 1.0 & {\cellcolor[HTML]{F2F2F2}} 1.0 & {\cellcolor[HTML]{D4DDE9}} 1.1 & {\cellcolor[HTML]{F2F2F2}} 1.0 & {\cellcolor[HTML]{F2F2F2}} 1.0 & {\cellcolor[HTML]{F2F2F2}} 0.2 & {\cellcolor[HTML]{E7ADBE}} 0.3 & {\cellcolor[HTML]{F2F2F2}} 0.2 & {\cellcolor[HTML]{F2F2F2}} 0.3 & {\cellcolor[HTML]{F2F2F2}} 0.2 & {\cellcolor[HTML]{D4DDE9}} 0.5 & {\cellcolor[HTML]{E7ADBE}} 0.2 & {\cellcolor[HTML]{E7ADBE}} 0.3 & {\cellcolor[HTML]{F2F2F2}} 1.7 & {\cellcolor[HTML]{F2F2F2}} 0.3 & {\cellcolor[HTML]{F2F2F2}} 0.4 \\

\textbf{MMG-orthogonal} & {\cellcolor[HTML]{F2F2F2}} 0.2 & {\cellcolor[HTML]{F2F2F2}} 0.4 & {\cellcolor[HTML]{F2F2F2}} 0.3 & {\cellcolor[HTML]{F2F2F2}} 0.4 & {\cellcolor[HTML]{F2F2F2}} 0.4 & {\cellcolor[HTML]{F2F2F2}} 0.4 & {\cellcolor[HTML]{F2F2F2}} 0.4 & {\cellcolor[HTML]{F2F2F2}} 1.0 & {\cellcolor[HTML]{F2F2F2}} 1.0 & {\cellcolor[HTML]{F2F2F2}} 1.0 & {\cellcolor[HTML]{F2F2F2}} 1.0 & {\cellcolor[HTML]{F2F2F2}} 0.3 & {\cellcolor[HTML]{F2F2F2}} 1.0 & {\cellcolor[HTML]{F2F2F2}} 1.3 & {\cellcolor[HTML]{F2F2F2}} 1.0 & {\cellcolor[HTML]{F2F2F2}} 0.4 & {\cellcolor[HTML]{F2F2F2}} 1.0 & {\cellcolor[HTML]{F2F2F2}} 0.6 & {\cellcolor[HTML]{F2F2F2}} 1.6 & {\cellcolor[HTML]{F2F2F2}} 0.4 & {\cellcolor[HTML]{F2F2F2}} 1.0 & {\cellcolor[HTML]{F2F2F2}} 1.0 & {\cellcolor[HTML]{F2F2F2}} 1.0 & {\cellcolor[HTML]{EED3DB}} 0.9 & {\cellcolor[HTML]{F2F2F2}} 1.0 & {\cellcolor[HTML]{F2F2F2}} 1.0 & {\cellcolor[HTML]{F2F2F2}} 1.0 & {\cellcolor[HTML]{F2F2F2}} 1.0 & {\cellcolor[HTML]{F2F2F2}} 1.0 & {\cellcolor[HTML]{E7ADBE}} 0.1 & {\cellcolor[HTML]{E7ADBE}} 0.3 & {\cellcolor[HTML]{F2F2F2}} 0.2 & {\cellcolor[HTML]{F2F2F2}} 0.3 & {\cellcolor[HTML]{F2F2F2}} 0.2 & {\cellcolor[HTML]{F2F2F2}} 0.4 & {\cellcolor[HTML]{F2F2F2}} 0.3 & {\cellcolor[HTML]{F2F2F2}} 0.4 & {\cellcolor[HTML]{F2F2F2}} 1.7 & {\cellcolor[HTML]{F2F2F2}} 0.3 & {\cellcolor[HTML]{F2F2F2}} 0.4 \\

\textbf{MMG-orthogonal-adaptive} & {\cellcolor[HTML]{F2F2F2}} 0.2 & {\cellcolor[HTML]{F2F2F2}} 0.4 & {\cellcolor[HTML]{F2F2F2}} 0.3 & {\cellcolor[HTML]{F2F2F2}} 0.4 & {\cellcolor[HTML]{F2F2F2}} 0.4 & {\cellcolor[HTML]{F2F2F2}} 0.4 & {\cellcolor[HTML]{F2F2F2}} 0.4 & {\cellcolor[HTML]{F2F2F2}} 1.0 & {\cellcolor[HTML]{F2F2F2}} 1.0 & {\cellcolor[HTML]{F2F2F2}} 1.0 & {\cellcolor[HTML]{F2F2F2}} 1.0 & {\cellcolor[HTML]{F2F2F2}} 0.3 & {\cellcolor[HTML]{F2F2F2}} 1.0 & {\cellcolor[HTML]{F2F2F2}} 1.3 & {\cellcolor[HTML]{F2F2F2}} 1.0 & {\cellcolor[HTML]{F2F2F2}} 0.4 & {\cellcolor[HTML]{F2F2F2}} 1.0 & {\cellcolor[HTML]{F2F2F2}} 0.6 & {\cellcolor[HTML]{F2F2F2}} 1.6 & {\cellcolor[HTML]{F2F2F2}} 0.4 & {\cellcolor[HTML]{F2F2F2}} 1.0 & {\cellcolor[HTML]{F2F2F2}} 1.0 & {\cellcolor[HTML]{F2F2F2}} 1.0 & {\cellcolor[HTML]{EED3DB}} 0.9 & {\cellcolor[HTML]{F2F2F2}} 1.0 & {\cellcolor[HTML]{F2F2F2}} 1.0 & {\cellcolor[HTML]{F2F2F2}} 1.0 & {\cellcolor[HTML]{F2F2F2}} 1.0 & {\cellcolor[HTML]{F2F2F2}} 1.0 & {\cellcolor[HTML]{F2F2F2}} 0.2 & {\cellcolor[HTML]{F2F2F2}} 0.4 & {\cellcolor[HTML]{F2F2F2}} 0.2 & {\cellcolor[HTML]{F2F2F2}} 0.3 & {\cellcolor[HTML]{F2F2F2}} 0.2 & {\cellcolor[HTML]{F2F2F2}} 0.4 & {\cellcolor[HTML]{F2F2F2}} 0.3 & {\cellcolor[HTML]{F2F2F2}} 0.4 & {\cellcolor[HTML]{F2F2F2}} 1.7 & {\cellcolor[HTML]{F2F2F2}} 0.3 & {\cellcolor[HTML]{F2F2F2}} 0.4 \\

\midrule
 & \begin{sideways} Asinh @ \textit{St} 1 × 1 (dof=1)\end{sideways} & 
\begin{sideways}Asinh @ \textit{St} 2 × 2 (dof=1) \end{sideways}& 
\begin{sideways}Asinh @ \textit{St} 3 × 3 (dof=2) \end{sideways}& 
\begin{sideways}Asinh @ \textit{St} 5 × 5 (dof=2)\end{sideways} &
\begin{sideways} Bimodal 1 × 1 \end{sideways}&
\begin{sideways}Bivariate \textit{Nm} 1 × 1\end{sideways} &
\begin{sideways} \textit{Hc} @ Bivariate \textit{Nm} 1 × 1\end{sideways} &
\begin{sideways} \textit{Hc} @ \textit{Mn} 25 × 25 (2-pair)\end{sideways} &
\begin{sideways} \textit{Hc} @ \textit{Mn} 3 × 3 (2-pair)\end{sideways} & 
\begin{sideways} \textit{Hc} @ \textit{Mn} 5 × 5 (2-pair) \end{sideways}& 
\begin{sideways} \textit{Mn} 2 × 2 (2-pair)\end{sideways} & 
\begin{sideways} \textit{Mn} 2 × 2 (dense)\end{sideways} &
\begin{sideways} \textit{Mn} 25 × 25 (2-pair)\end{sideways} & 
\begin{sideways} \textit{Mn} 25 × 25 (dense) \end{sideways}& 
\begin{sideways} \textit{Mn} 3 × 3 (2-pair) \end{sideways}& 
\begin{sideways} \textit{Mn} 3 × 3 (dense) \end{sideways}&
\begin{sideways} \textit{Mn} 5 × 5 (2-pair) \end{sideways}& 
\begin{sideways} \textit{Mn} 5 × 5 (dense) \end{sideways}&
\begin{sideways} \textit{Mn} 50 × 50 (dense) \end{sideways}& 
\begin{sideways} \textit{Nm} CDF @ Bivariate \textit{Nm} 1 × 1\end{sideways} &
\begin{sideways} \textit{Nm} CDF @ \textit{Mn} 25 × 25 (2-pair) \end{sideways}&
\begin{sideways} \textit{Nm} CDF @ \textit{Mn} 3 × 3 (2-pair) \end{sideways}& 
\begin{sideways} \textit{Nm} CDF @ \textit{Mn} 5 × 5 (2-pair) \end{sideways}& 
\begin{sideways} \textit{Sp} @ \textit{Mn} 25 × 25 (2-pair) \end{sideways}& 
\begin{sideways} \textit{Sp} @ \textit{Mn} 3 × 3 (2-pair)\end{sideways} & 
\begin{sideways} \textit{Sp} @ \textit{Mn} 5 × 5 (2-pair)\end{sideways} & 
\begin{sideways} \textit{Sp} @ \textit{Nm} CDF @ \textit{Mn} 25 × 25 (2-pair) \end{sideways}&
\begin{sideways} \textit{Sp} @ \textit{Nm} CDF @ \textit{Mn} 3 × 3 (2-pair) \end{sideways}& 
\begin{sideways} \textit{Sp} @ \textit{Nm} CDF @ \textit{Mn} 5 × 5 (2-pair) \end{sideways}&
\begin{sideways} \textit{St} 1 × 1 (dof=1)\end{sideways} & 
\begin{sideways} \textit{St} 2 × 2 (dof=1)\end{sideways} &
\begin{sideways} \textit{St} 2 × 2 (dof=2) \end{sideways}& 
\begin{sideways} \textit{St} 3 × 3 (dof=2)\end{sideways} & 
\begin{sideways} \textit{St} 3 × 3 (dof=3)\end{sideways} & 
\begin{sideways} \textit{St} 5 × 5 (dof=2)\end{sideways} & 
\begin{sideways} \textit{St} 5 × 5 (dof=3) \end{sideways}& 
\begin{sideways} Swiss roll 2 × 1 \end{sideways}&
\begin{sideways} Uniform 1 × 1 (additive noise=0.1) \end{sideways}&
\begin{sideways} Uniform 1 × 1 (additive noise=0.75) \end{sideways}& 
\begin{sideways} Wiggly @ Bivariate \textit{Nm} 1 × 1 \end{sideways} \\

\bottomrule
\end{tabular}}

\caption{Low MI estimation over 10 seeds using N = 10k test samples against ground truth (GT)~\citep{czyż2023normalevaluationmutualinformation}. All methods were trained with 100k samples. \textbf{Color indicates relative negative (red) and positive bias (blue)}.  Blank entries indicate that an estimator experienced numerical instabilities. List of abbreviations ( \textit{Mn}: Multinormal,  \textit{St}: Student-t, \textit{Nm}: Normal, \textit{Hc}: Half-cube, \textit{Sp}: Spiral)}

\label{tab:benchmark}
\end{table}


\subsection{High MI Benchmark}
\begin{figure}[ht]
    \centering
    \includegraphics[width=\textwidth]{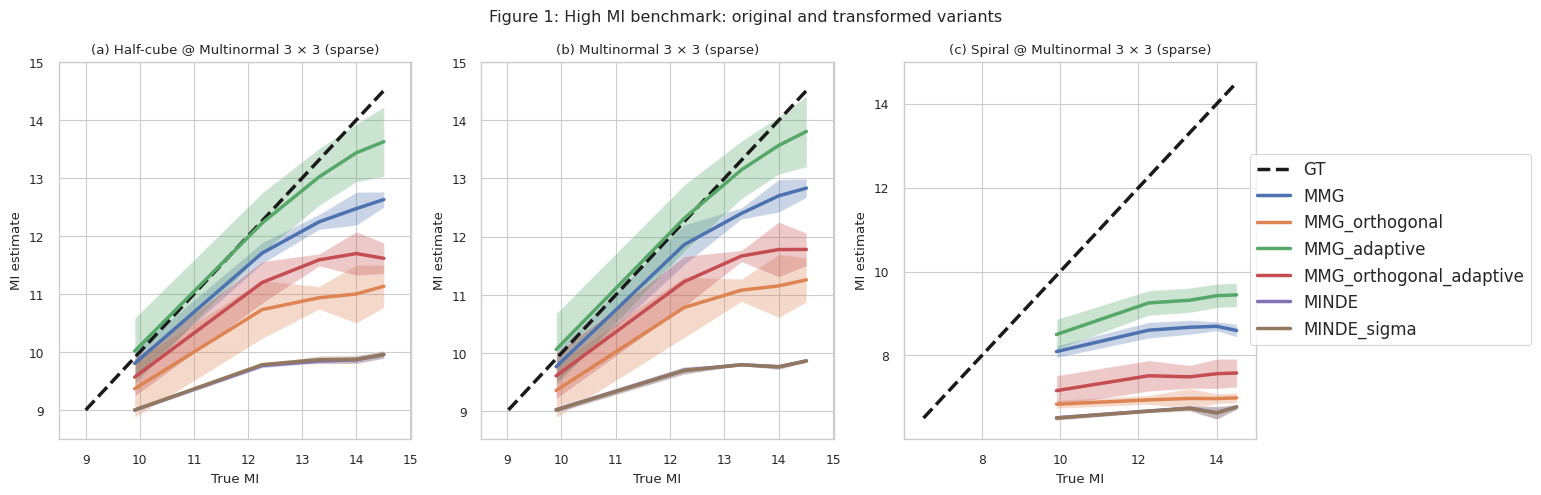}
    \caption{High MI benchmark: original (column (b)) and transformed variants (columns (a) and (c)).}
    \label{fig:hm}
\end{figure}

To test the robustness and limits of our estimator, we extend the high-MI study from MINDE \citep{minde}, pushing their experimental setup from a range of MI $\leq$ 5 into the significantly more challenging regime of MI $\in [10, 15]$. Following their protocol, we use a sparse $3 \times 3$ Multinormal setup and its two MI-preserving transforms (Half-cube, Spiral). Figure~\ref{fig:hm} reports the mean $\pm$ one standard deviation over 10 seeds.

On the \textit{original} and \textit{Half-cube} cases (Fig.~\ref{fig:hm}a\&b), \textbf{MMG-adaptive} remains the most accurate as MI grows, while other estimators falter due to distinct sources of error. \textbf{MINDE} struggles significantly as its score-matching objective requires approximating the sharp, high-frequency score functions of high-MI distributions, a task where neural networks' spectral bias leads to significant underestimation~\cite{rahaman2019spectral}. In contrast, our orthogonal variants exhibit a different limitation: a systematic conservative bias. This occurs because the orthogonal estimator measures the distance between neural network approximations of the denoisers, and the distance between these "smoothed-out" approximations is inherently smaller than the true distance between the optimal denoisers. This underestimation bias becomes more pronounced as the true MI gap grows larger.

This reveals a clear bias-variance trade-off dictating the optimal estimator. For the broad low-MI benchmark, the main challenge is variance, making the stable \textbf{MMG-orthogonal-adaptive} the superior choice. In this high-MI setting, however, this systematic bias becomes the dominant error, rendering the less-biased \textbf{MMG-adaptive} more accurate. This relative performance ranking holds even on the highly non-linear \textit{Spiral} transform (Fig.~\ref{fig:hm}c), where all methods are challenged.

\subsection{Consistency Test}

\begin{figure}[ht]
    \centering
        
    \begin{subfigure}[b]{0.31\textwidth}
        \includegraphics[width=\linewidth]{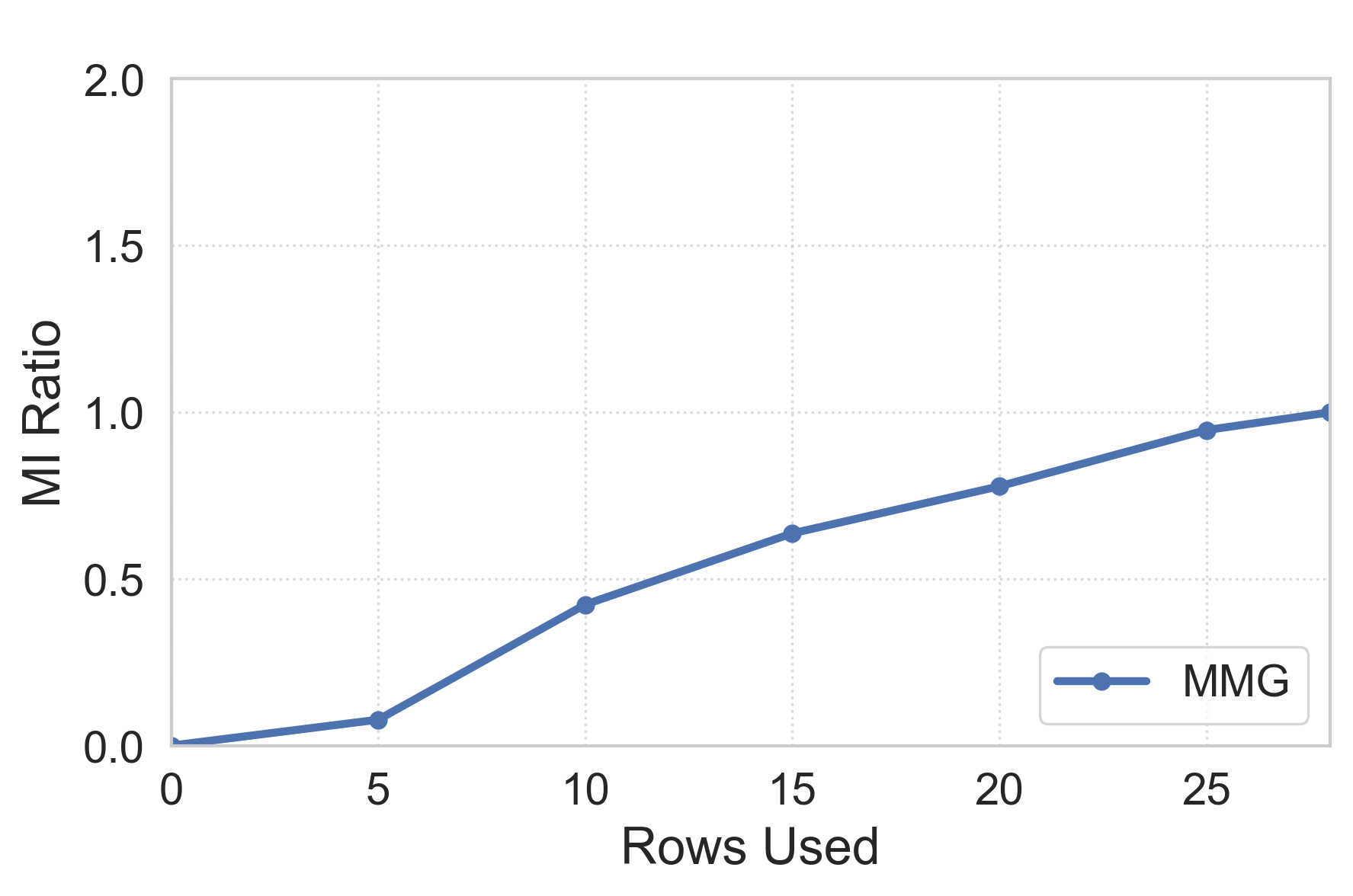}
        \caption{Baseline test}
        \label{fig:set_1}
    \end{subfigure}
    \begin{subfigure}[b]{0.31\textwidth}
        \includegraphics[width=\linewidth]{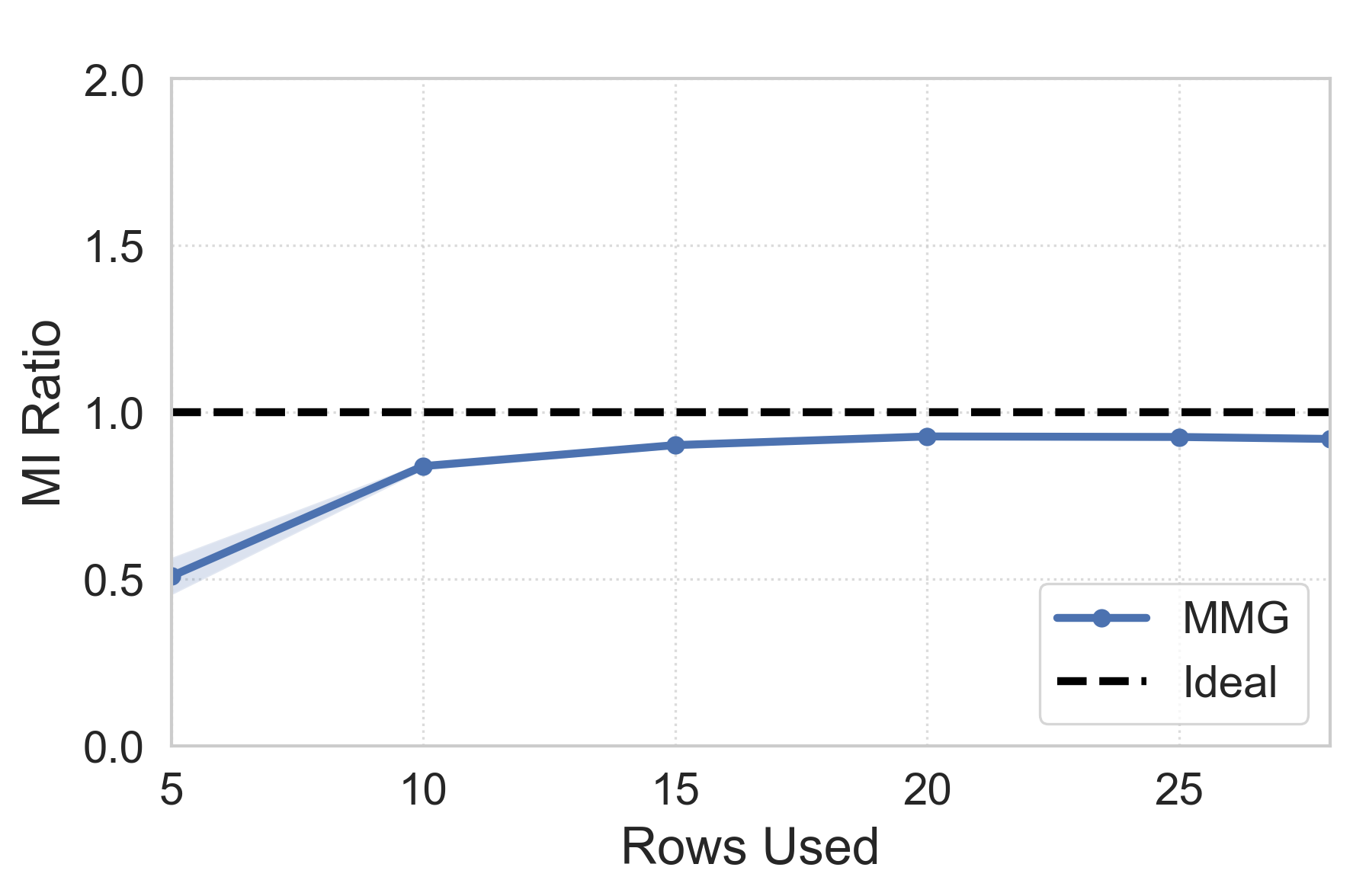}
        \caption{Data processing test}
        \label{fig:set_2}
    \end{subfigure}
    \begin{subfigure}[b]{0.31\textwidth}
        \includegraphics[width=\linewidth]{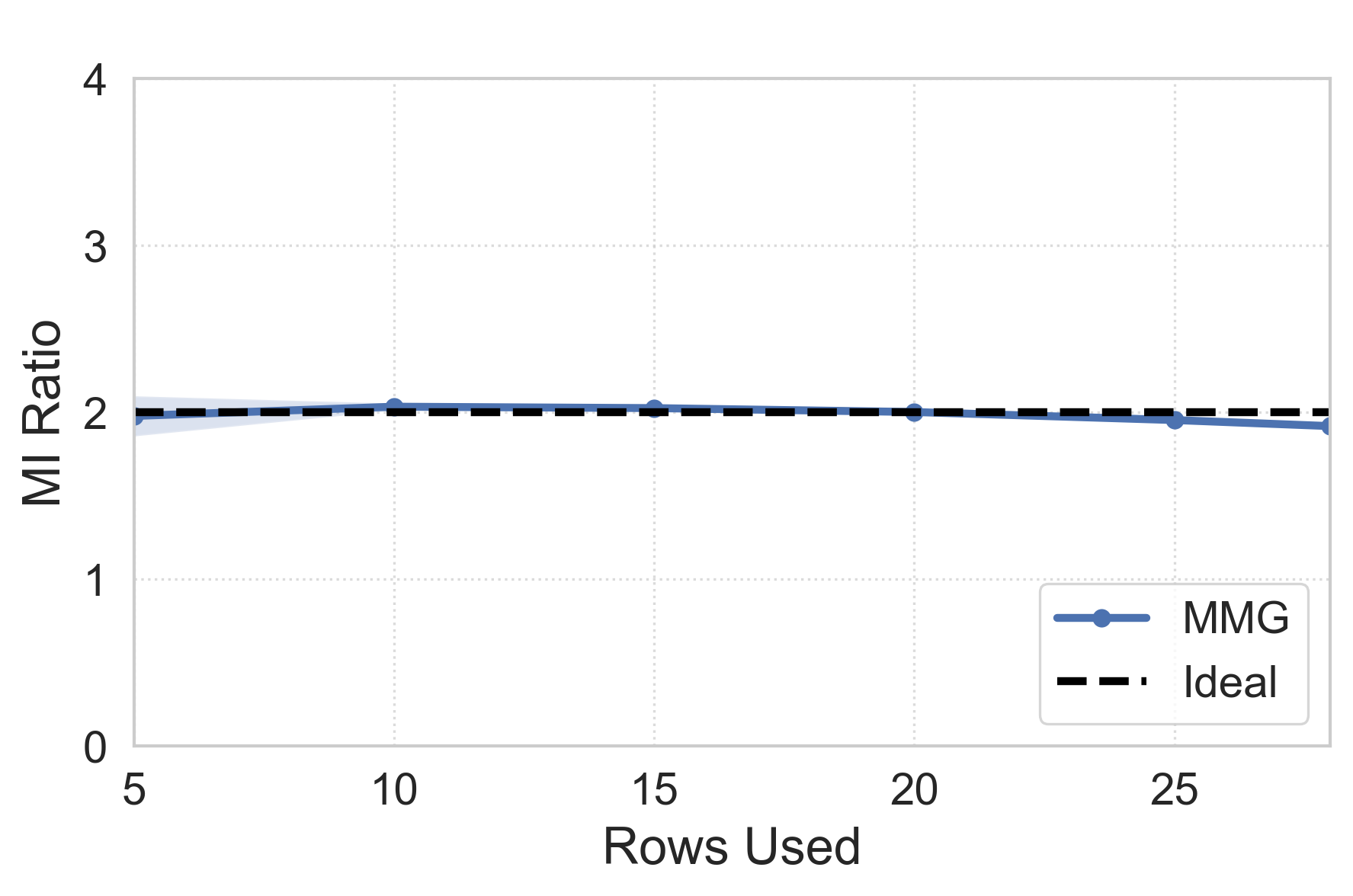}
        \caption{Additivity test}
        \label{fig:set_3}
    \end{subfigure}

    \caption{Consistency Tests over MNIST dataset: (a) evaluation of $\frac{I(A; B_r)}{I(A; B)}$; (b) evaluation of $\frac{I(A; [B_{r+k}, B_r])}{I(A; B_{r+k})}$ for $k>0$; (c) evaluation of $\frac{I([A^1, A^2]; [B^1_r, B^2_r])}{I(A^1; B^1_r)}$.
    }
    \label{fig:consistency_test}
\end{figure}

We conduct self-consistency tests inspired by \citet{song2019understanding} to evaluate the properties of MMG using high-dimensional real-world data, specifically samples from the MNIST dataset (28×28 resolution)~\citep{mnist}. Let $A$ represent an image and $B_r$ denote the image consisting of the top $r$ rows of $A$. These tests are designed to verify whether the estimators adhere to fundamental properties of MI through three subtasks: Baseline Test, Data-Processing Test, and Additivity Test for two independent images sampled from dataset. Ideally, as $r$ increases, $\frac{I(A; B_r)}{I(A; B)}$ should monotonically approach 1, $\frac{I(A; [B_{r+k}, B_r])}{I(A; B_{r+k})}$ should consistently equal 1, and $\frac{I([A^1, A^2]; [B^1_r, B^2_r])}{I(A^1; B^1_r)}$ should consistently equal 2. To ensure a fair comparison with MINDE~\citep{minde}, we aligned our experimental settings and parameters and tested MMG using five random seeds. The results of the three tests are presented in Figure~\ref{fig:consistency_test}. Overall, MMG performed well and successfully passed all tests.

\section{Related Work}

Estimating mutual information (MI) from samples is a central challenge in machine learning. Traditional non-parametric approaches, such as methods based on data binning or kernel density estimation (KDE), struggle with the curse of dimensionality. More advanced estimators based on k-nearest neighbors have shown significant improvements \citep{kraskov2004, pal2010estimation}, but can still be challenged by complex, high-dimensional data \citep{gao2015efficient}. Consequently, the modern paradigm is dominated by neural network-based approaches that optimize a variational bound on the MI.

\paragraph{Variational Bounds on Mutual Information.}
Most recent neural MI estimators are built upon variational lower bounds derived from f-divergences. The pioneering work, MINE \citep{belghazi2018mutual}, leverages the Donsker-Varadhan representation of the KL-divergence, spurring related estimators like the variance-reduced SMILE \citep{song2019understanding} and DoE \citep{mcallester2020formal}. A particularly successful family is based on contrastive learning, where estimators like InfoNCE \citep{oord2018representation} train a critic to distinguish between joint and marginal samples \citep{poole2019variational}. However, these methods face significant limitations: their sample complexity can scale exponentially with the true MI, often being capped by the logarithm of the batch size \citep{mcallester2020formal}. This difficulty can be viewed as a "density chasm": the marginal distribution $p(x)p(y)$ is often a poor proposal for the joint $p(x,y)$, leading to high-variance estimates, especially in high-MI settings \citep{rhodes2020telescoping, brekelmans2021improving}.

\paragraph{Mutual Information Estimation with Diffusion Models.}
Denoising diffusion models offer a natural and powerful framework to bridge this density chasm. They define a continuous process that transforms a complex data distribution into a simple tractable one, providing a path of intermediate distributions. The potential of this framework for MI estimation was recently demonstrated by \textbf{MINDE} \citep{minde}, which connects MI to the difference between conditional and unconditional score functions ($\nabla_{\vx} \log p(\vx)$). Our work, \textbf{MMG}, builds on a different and more direct connection \citep{guo, itd}. Instead of relying on learned score functions, we show that MI corresponds exactly to the integrated gap between the Minimum Mean Square Error (MMSE) of conditional and unconditional denoising. This formulation connects MI directly to the denoising objective itself, rather than its gradient, providing an elegant and potentially more robust pathway for estimation.

\section{Conclusion}
In this work, we introduced MMG, a principled and robust estimator for mutual information derived from the information-theoretic properties of denoising diffusion models. Our method directly connects MI to the integrated Minimum Mean Square Error (MMSE) gap between a conditional and an unconditional denoiser. We further enhanced this framework with two key techniques: an adaptive importance sampling scheme to improve integration accuracy and an orthogonal principle to increase estimator stability.

Through extensive experiments, we demonstrated that MMG achieves exceptional accuracy and robustness, successfully providing stable estimates on 39 out of 40 tasks in a comprehensive benchmark, and successfully passes all self-consistency tests. Notably, our method excels in the challenging high-MI regime, significantly outperforming current score-based diffusion estimators. Our analysis also uncovered a fundamental bias-variance trade-off, revealing that the optimal estimator configuration—either with or without the orthogonal principle—depends on the MI magnitude of the problem. Future work could explore strategies to automatically navigate this bias-variance trade-off or apply the MMSE-gap framework to other information-theoretic quantities.

\bibliography{neurips_2025.bib} 
\bibliographystyle{iclr2025_conference}

\appendix

\section{Derivation of the Orthogonal Principle}\label{sec:ortho_proof}

This section provides a brief derivation for the orthogonal principle (Equation~\ref{eq:ortho}) used in our estimator, following the original work of \cite{kong2023interpretable}.

The principle is based on the following identity. Let $\hat{\vx}(\vz) \equiv \hat{\vx} = \mathbb{E}[\vx|\vz]$ be the unconditional MMSE denoiser and $\hat{\vx}(\vz, y) \equiv \hat{\vx}_y = \mathbb{E}[\vx|\vz, y]$ be the conditional one. The identity states:
\begin{equation}\label{eq:ortho_appendix}
\mathbb{E}[\|\vx - \hat{\vx}\|^2] - \mathbb{E}[\|\vx - \hat{\vx}_y\|^2] = \mathbb{E}[\|\hat{\vx}_y - \hat{\vx}\|^2]
\end{equation}
Below is a brief proof sketch for Equation~\ref{eq:ortho_appendix}.

\paragraph{Proof Sketch.} The proof relies on the law of total expectation and the orthogonality property of MMSE estimation (i.e., the estimation error is orthogonal to any function of the conditioning variables). We begin by showing a key cross-term is zero:
\begin{align*}
    \mathbb{E}[(\vx - \hat{\vx}_y) \cdot (\hat{\vx}_y - \hat{\vx})] &= \mathbb{E}\left[ \mathbb{E}[(\vx - \hat{\vx}_y) \cdot (\hat{\vx}_y - \hat{\vx}) | \vz, y] \right] & \text{(Law of Total Expectation)} \\
    &= \mathbb{E}\left[ (\mathbb{E}[\vx | \vz, y] - \hat{\vx}_y) \cdot (\hat{\vx}_y - \hat{\vx}) \right] & \text{(Pulling out known terms)} \\
    &= \mathbb{E}\left[ (\hat{\vx}_y - \hat{\vx}_y) \cdot (\hat{\vx}_y - \hat{\vx}) \right] = 0 & \text{(Definition of $\hat{\vx}_y$)}
\end{align*}
Because this cross-term is zero, we can expand the unconditional error $\|\vx - \hat{\vx}\|^2$ as follows:
\begin{align*}
\mathbb{E}[\|\vx - \hat{\vx}\|^2] &= \mathbb{E}[\|(\vx - \hat{\vx}_y) + (\hat{\vx}_y - \hat{\vx})\|^2] \\
&= \mathbb{E}[\|\vx - \hat{\vx}_y\|^2] + \mathbb{E}[\|\hat{\vx}_y - \hat{\vx}\|^2] + 2 \cdot \underbrace{\mathbb{E}[(\vx - \hat{\vx}_y) \cdot (\hat{\vx}_y - \hat{\vx})]}_{0} \\
&= \mathbb{E}[\|\vx - \hat{\vx}_y\|^2] + \mathbb{E}[\|\hat{\vx}_y - \hat{\vx}\|^2]
\end{align*}
Rearranging the terms yields the identity in Equation~\ref{eq:ortho_appendix}.

\renewcommand{\tabcolsep}{2.0pt}

\begin{table}[h]
\caption{MMG network training hyper-parameters. $Dim$ of the task correspond the sum of the two variables dimensions, and $d$ corresponds to the randomization probability.}
\centering
\begin{tabular}{ccccccccc}
\toprule
Benchmark $Dim$ & $d$ & Width & Time Embedding Size & Batch Size & $lr$ & Iterations & \# of Params \\
\midrule
$\leq$ 10 & 0.5 & 64 & 64 & 128 & 1e-3 & 390k & 55425 \\
50 & 0.5 & 128 & 128 & 256 & 2e-3 & 290k & 220810 \\
100 & 0.5 & 256 & 256 & 256 & 2e-3 & 290k & 898354 \\
\midrule
Consistency Tests & 0.5 & 256 & 256 & 64 & 1e-3 & 390k & 1597968 \\
\bottomrule
\end{tabular}
\label{table:mind}
\end{table}

\begin{table}[h]
\centering
\begin{tabular}{ccccccccccc}
\toprule
Sampling Parameter & LogSNR Loc & LogSNR Scale & Clip & EMA Decay & Inference Times& N\_Points \\
\midrule
- & 2.0 & 3.0 & 4.0 & 0.999 & 10 & 10000 \\
\bottomrule
\end{tabular}
\caption{Sampling Hyperparameters}
\label{table:fixed_params}
\end{table}

\section{Ablation Study on Adaptive Sampling}

In this section, we conduct an ablation study to specifically evaluate the impact of our adaptive importance sampling strategy. We compare the performance of two non-orthogonal estimator variants:
\begin{itemize}
    \item \textbf{MMG-adaptive ("Adaptive"):} Employs the adaptive importance sampling scheme described in Section~\ref{sec:implementation}.
    \item \textbf{MMG ("Baseline"):} Uses a fixed, default importance sampling distribution. 
\end{itemize}
This comparison is designed to isolate the effect of the sampling strategy on estimation accuracy and stability, particularly across a diverse set of distributions. The results are presented in Table~\ref{tab:ablation_sampling}.

\begin{table}[htbp]
\centering
\caption{Ablation study comparing MMG-adaptive ("Adaptive") with adaptive importance sampling against the baseline MMG ("Baseline") with a fixed sampling distribution. This analysis uses the non-orthogonal variants to isolate the impact of the sampling strategy. For each task, the estimate closer to the Ground Truth and the lower standard deviation (Std) are independently bolded.} 
\label{tab:ablation_sampling}
\begin{tabular}{lccccc}
\toprule
\multirow{2}{*}{\textbf{Task}} & \multirow{2}{*}{\textbf{Ground Truth}} & \multicolumn{2}{c}{\textbf{Adaptive}} & \multicolumn{2}{c}{\textbf{Baseline}} \\ 
\cmidrule(lr){3-4} \cmidrule(lr){5-6}
& & \textbf{Estimate} & \textbf{Std} & \textbf{Estimate} & \textbf{Std} \\
\midrule
\multicolumn{6}{l}{\textit{Basic Distribution Tasks}} \\
1v1-normal-0.75 & 0.4133 & \textbf{0.4160} & \textbf{0.0447} & 0.4208 & 0.0754 \\
1v1-additive-0.1 & 1.7094 & 1.6929 & \textbf{0.0494} & \textbf{1.6984} & 0.0677 \\
1v1-bimodal-0.75 & 0.4133 & \textbf{0.4133} & \textbf{0.0612} & 0.4201 & 0.0766 \\
\hline
\multicolumn{6}{l}{\textit{High Dimensional Tasks}} \\
multinormal-dense-25-25-0.5 & 1.2922 & \textbf{1.2063} & \textbf{0.1636} & 1.1603 & 0.2526 \\
multinormal-dense-50-50-0.5 & 1.6243 & \textbf{1.7926} & \textbf{0.3053} & 1.8493 & 0.4398 \\
\hline
\multicolumn{6}{l}{\textit{Non-Gaussian Distribution Tasks}} \\
student-identity-1-1-1 & 0.2242 & 0.3644 & \textbf{0.2472} & \textbf{0.3583} & 0.2884 \\
student-identity-3-3-2 & 0.2909 & \textbf{0.4901} & \textbf{0.6502} & 0.5123 & 0.7055 \\
student-identity-5-5-2 & 0.4482 & 0.7747 & 1.0550 & \textbf{0.7683} & \textbf{1.0492} \\
\hline
\multicolumn{6}{l}{\textit{Complex Transformation Tasks}} \\
spiral-multinormal-sparse-3-3-2.0 & 1.0217 & \textbf{1.0012} & \textbf{0.0805} & 1.0152 & 0.1187 \\
spiral-multinormal-sparse-25-25-2.0 & 1.0217 & \textbf{0.8713} & \textbf{0.1557} & 0.8215 & 0.2683 \\
\bottomrule
\end{tabular}
\end{table}

The results in Table~\ref{tab:ablation_sampling} demonstrate the consistent benefits of the adaptive sampling strategy. The "Adaptive" method frequently achieves a lower standard deviation, indicating improved estimator stability. This is particularly evident in high-dimensional and complex transformation tasks, such as `multinormal-dense-50-50-0.5` and `spiral-multinormal-sparse-25-25-2.0`, where the reduction in variance is substantial. While the accuracy of the point estimate is competitive across both methods, the adaptive approach often provides estimates closer to the ground truth in the more challenging settings. Overall, this ablation study validates that tailoring the sampling distribution to the specific data leads to a more robust and reliable MI estimator.

\section{Implementation Details}
We follow the implementation of \cite{franzese2023minde} which uses stacked multi-layer perception (MLP) with skip connections. We adopt a simplified version of the same  network architecture: this involves three Residual MLP blocks. We use the \textit{Adam optimizer} \citep{kingma2014adam} for training and Exponential moving average (EMA) with a momentum parameter $m = 0.999$. We use the \textit{ReduceLROnPlateau} scheduler with a patience of 200 epochs, reducing the learning rate by half if the training loss does not improve after 200 epochs. We returned the mean estimate on the test data set over 10 runs. All experiments are run on NVIDIA RTX A6000 GPUs.
The hyper-parameters are presented in Table \ref{table:mind} for  MMG. Concerning the consistency tests, we independently train an autoencoder for each version of the \textsc{mnist} dataset with $r$ rows available.

\section{Visual Analysis of the MMG Estimator}

This section provides an intuitive, qualitative analysis of the MMG estimator's integrand to support the core components of our method. By visualizing the integrand on a challenging three-dimensional, Spiral-transformed sparse Multinormal distribution (GT MI = 9.90), we can clearly see the complementary roles of the orthogonal principle for stability and adaptive importance sampling for accuracy. The plots below were generated by densely sampling 10,000 log SNR points and then binning the results (bin=50) to illustrate the underlying trends.

\begin{figure}[H]
\centering
\begin{subfigure}[b]{0.49\textwidth}
\centering
\includegraphics[width=\textwidth]{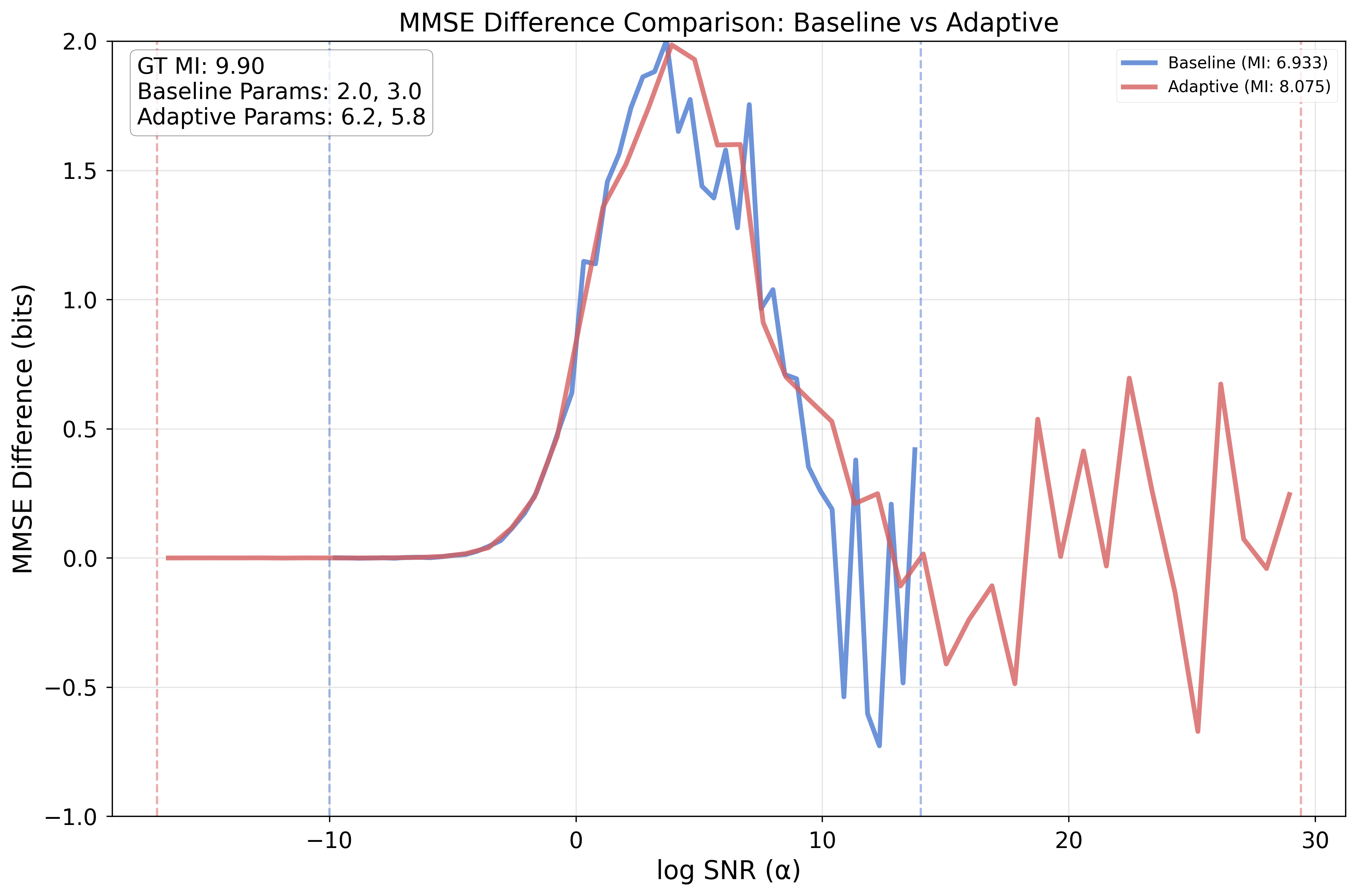}
\caption{MMSE Difference}
\label{fig:mmse_diff}
\end{subfigure}
\hfill
\begin{subfigure}[b]{0.49\textwidth}
\centering
\includegraphics[width=\textwidth]{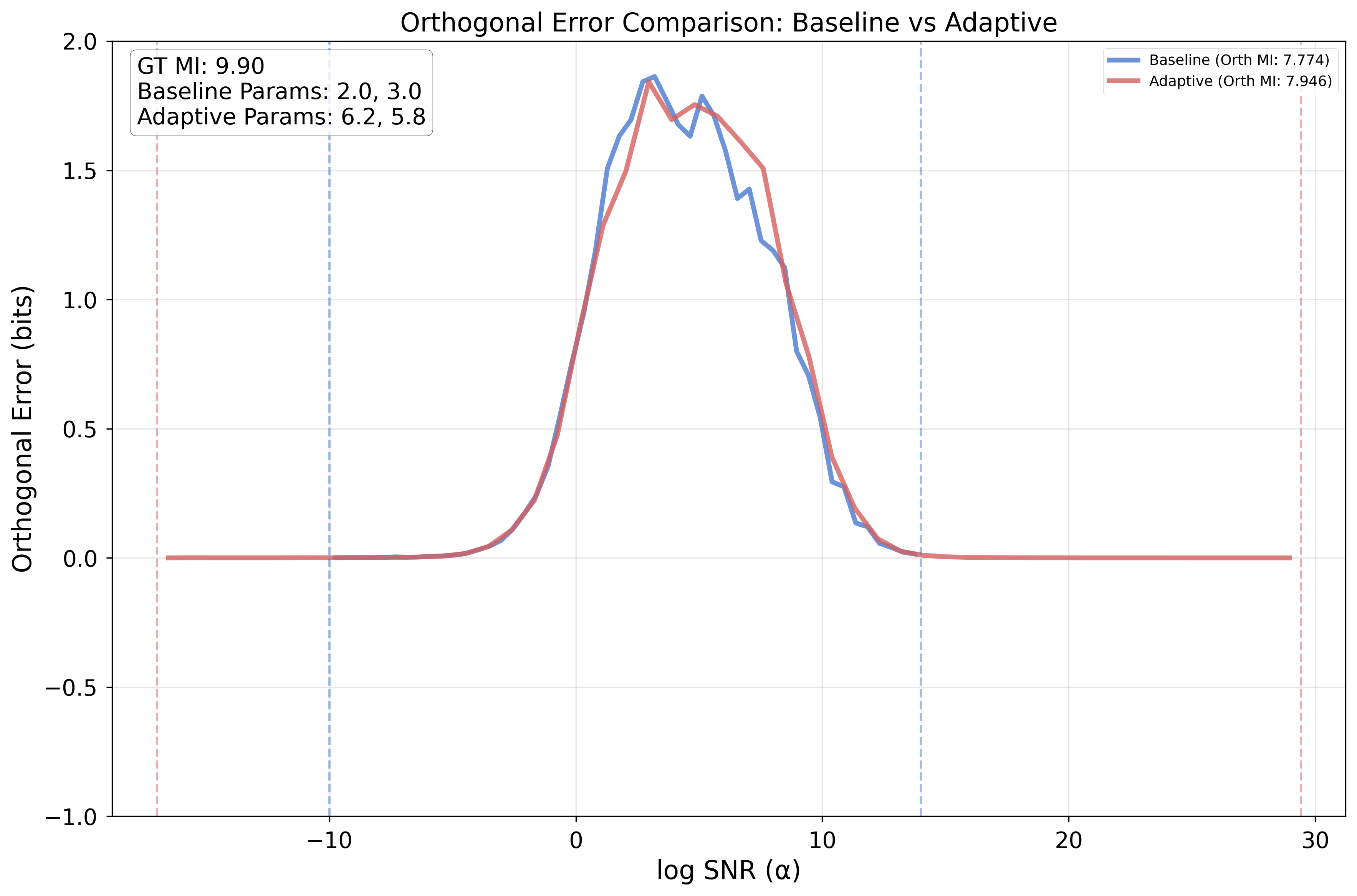}
\caption{Orthogonal Error}
\label{fig:orth_err}
\end{subfigure}
\caption{\textbf{Integrand analysis on a Spiral-transformed task (GT MI = 9.90).} Comparison between (a) the volatile direct MMSE subtraction method and (b) the stable orthogonal principle. In both plots, the adaptive sampler (red) outperforms the baseline (blue) by focusing on the most informative region.}
\label{fig:appendix_analysis}
\end{figure}

Figure~\ref{fig:appendix_analysis} directly compares the integrand calculated via direct MMSE subtraction against the one derived from our orthogonal principle.

\paragraph{Analysis of the Orthogonal Principle's Contribution}
The orthogonal principle's primary contribution is enhancing integrand stability. As shown in Figure~\ref{fig:orth_err}, it replaces the highly volatile direct subtraction method with an exceptionally smooth and non-negative integrand, crucial for reliable numerical integration. This stability, however, introduces a trade-off that becomes particularly evident in \textbf{high mutual information estimation}: the orthogonal integrand's peak is lower, which can lead to a conservative bias by underestimating the true distance between optimal denoisers. While the dramatic reduction in variance makes it a more robust choice for general cases, this systematic underestimation can become a limiting factor when the true MI is large.

\paragraph{Analysis of Adaptive Sampling's Contribution}
Adaptive sampling boosts estimation accuracy and efficiency. As illustrated in Figure~\ref{fig:orth_err}, while a fixed sampler may be misaligned with the integrand's peak, our adaptive method dynamically concentrates samples in this most informative SNR region. This targeted strategy leads to a demonstrably more accurate MI estimate (\textbf{8.075 bits} vs. \textbf{6.933 bits} for the direct method), confirming the value of focusing the integration where it matters most.

\end{document}